\documentclass[letterpaper, 10 pt, conference]{ieeeconf}  

\IEEEoverridecommandlockouts                              

\usepackage{bm}
\usepackage{url}
\usepackage{graphicx}
\usepackage{lipsum}
\usepackage{pifont}
\usepackage{amsmath}
\usepackage{amssymb}
\usepackage{nccmath}
\usepackage{comment}
\usepackage{balance}
\usepackage{multirow}
\usepackage{textcomp}
\usepackage{multirow}
\usepackage{subcaption}
\usepackage{booktabs}
\usepackage{mathtools}
\usepackage{algorithm}
\usepackage{dblfloatfix}
\usepackage[table]{xcolor}
\usepackage[noend]{algpseudocode}

\DeclareMathOperator*{\argmin}{arg\,min}

\title{\LARGE \bf
MegaParticles: Range-based 6-DoF Monte Carlo Localization \\
with GPU-Accelerated Stein Particle Filter
}

\author{Kenji Koide$^{1}$, Shuji Oishi$^{1}$, Masashi Yokozuka$^{1}$, and Atsuhiko Banno$^{1}$
\thanks{*This work was supported in part by JSPS KAKENHI Grant Number 23K16979 and a project commissioned by the New Energy and Industrial Technology Development Organization (NEDO).}
\thanks{$^{1}$All the authors are with the Department of Information Technology and Human Factors, the National Institute of Advanced Industrial Science and Technology, Tsukuba, Ibaraki, Japan, {\tt\small k.koide@aist.go.jp}}%
}

\begin{document}

\maketitle
\thispagestyle{empty}
\pagestyle{empty}

\setlength\floatsep{8pt}
\setlength\textfloatsep{8pt}

\begin{abstract}

This paper presents a 6-DoF range-based Monte Carlo localization method with a GPU-accelerated Stein particle filter. To update a massive amount of particles, we propose a Gauss-Newton-based Stein variational gradient descent (SVGD) with iterative neighbor particle search. This method uses SVGD to collectively update particle states with gradient and neighborhood information, which provides efficient particle sampling. For an efficient neighbor particle search, it uses locality sensitive hashing and iteratively updates the neighbor list of each particle over time. The neighbor list is then used to propagate the posterior probabilities of particles over the neighbor particle graph. The proposed method is capable of evaluating one million particles in real-time on a single GPU and enables robust pose initialization and re-localization without an initial pose estimate. In experiments, the proposed method showed an extreme robustness to complete sensor occlusion (i.e., kidnapping), and enabled pinpoint sensor localization without any prior information.

\end{abstract}

\section{Introduction}

Reliable sensor localization is crucial for autonomous systems such as service robots and autonomous driving vehicles. In particular, point-cloud-based localization has been widely used in many applications as a result of the emergence of precise and affordable range sensors. Although the recent development of scan-matching-based localization techniques (e.g., sliding window optimization \cite{Koide_2022} and the tight coupling of LiDAR and IMU constraints \cite{Xu2022}) has significantly improved the pose tracking accuracy and reliability, it is still challenging to deal with cases where no prior knowledge of the sensor position is available (e.g., initial position estimation without GNSS and re-localization after kidnapping).

Monte Carlo localization (MCL) is a category of localization methods that estimate the sensor pose using random-sampling-based probabilistic inference \cite{probrobo}. In particular, a particle filter is the most common approach used for 2D LiDAR localization and mapping \cite{Fox_2003, Grisetti_2005, Kuang_2023}. Owing to its non-linear, non-Gaussian nature, a particle filter enables reliable localization even in cases where single hypothesis algorithms suffer from observational ambiguity and repeated environmental structures. However, because the number of particles required to fill the state space grows exponentially as the dimension increases, it has generally been considered to be difficult to apply the Monte Carlo approach to 6-DoF pose estimation in 3D environments.

Several studies have tackled 6-DoF MCL with carefully designed sampling techniques (e.g., RBPF with rotation translation decoupling \cite{Deng_2021}). While these methods enable real-time 6-DoF pose estimation with fewer particles, it is still difficult to deal with situations where strong ambiguity causes state distributions with many modes. The recently proposed Stein particle filter \cite{Maken_2022} combines a particle filter with Stein variational gradient descent (SVGD) \cite{NIPS2016_b3ba8f1b}. SVGD updates the particle states {\it collectively} to improve the sampling efficiency with gradient information while retaining the diversity of the particles. Although a Stein particle filter has been shown to enable 6-DoF pose estimation in a 3D environment with only 50 particles, it requires a costly kernel-based computation that prevents increasing the number of particles.

\begin{figure}[tb]
  \centering
  \includegraphics[width=\linewidth]{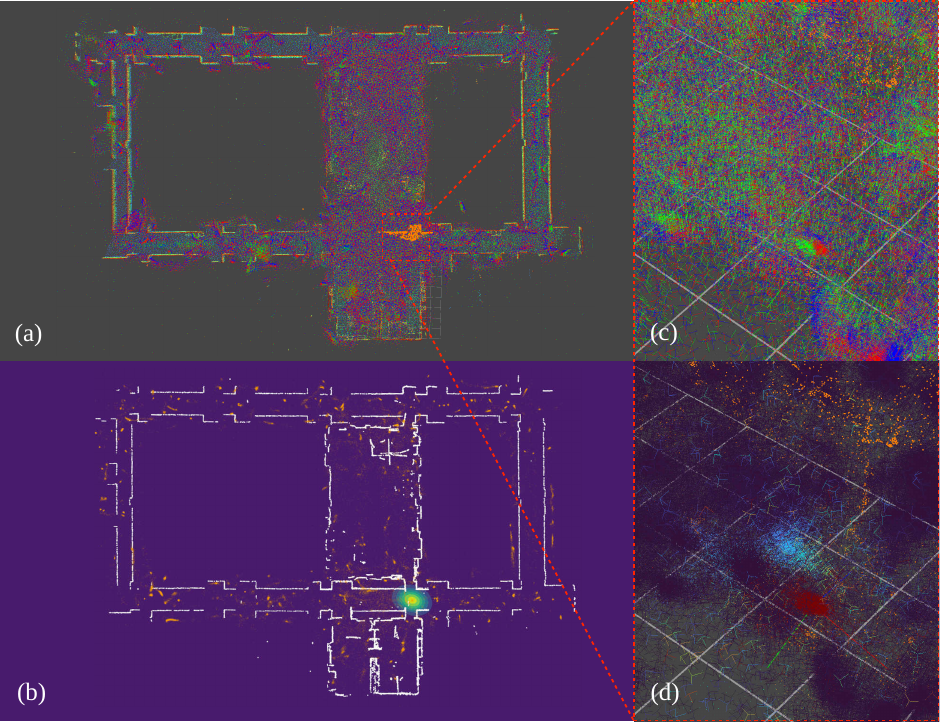}
  \caption{(a) Proposed method performs 6-DoF sensor localization with one million particles. All the particles are evaluated and updated in real-time on a single GPU. Point clouds acquired by a MS Azure Kinect are used (No IMU input). (b) Posterior probability distribution. (c) A close look at the maximum posterior particle (zoom in to see RGB-colored pose particles), and (d) the same view with particles colored based on the posterior probabilities. }
  \label{fig:ampap}
\end{figure}

In this work, we propose a 6-DoF Monte Carlo localization method that combines GPU parallel processing and SVGD-based particle state optimization \footnote{See the project page for supplementary videos: \url{https://staff.aist.go.jp/k.koide/projects/icra2024_mp/}}. The proposed method was designed to fully leverage the GPU acceleration, and it is capable of evaluating and updating a million particles ($1024^2$ particles) in real-time, as shown in Fig. \ref{fig:ampap}. Inspired by \cite{Maken_2022}, we use SVGD \cite{NIPS2016_b3ba8f1b} to update the particle states for efficient sampling. To efficiently perform the SVGD-based collective particle state optimization, we propose an approximated Gauss-Newton SVGD with locality sensitive hashing (LSH)-based neighbor particle search in the SE3 pose space. Furthermore, to quickly estimate and propagate the particle posterior probabilities, we developed a Bayesian filter on a dynamic neighbor particle graph.

To demonstrate the robustness and re-localization ability of the proposed method, we conducted experiments in an indoor environment with repeated and symmetric structures and in a large outdoor environment with dynamic objects and vegetation changes. The experimental results showed that the proposed method could robustly estimate the sensor pose starting from a particle set initialized with a uniform 6-DoF pose distribution. It was also robust to the complete blockage of sensor data (i.e., kidnapping), which is difficult, if not impossible, to overcome with existing scan-matching-based localization methods. Furthermore, the neighbor-particle-graph-based posterior probability estimation enabled pinpoint localization once the sensor moved and the pose ambiguity was resolved.

\section{Related Work}

\subsection{Iterative Scan Matching-based Localization}

The most common approach to 3D map localization is to iteratively apply point cloud scan matching (e.g., ICP \cite{Chetverikova} and NDT \cite{magnusson2009three}) between LiDAR scan and map points. Because these scan matching methods require a precise initial guess for convergence and are error-prone in feature-less environments, they are often integrated with other data sources (e.g., wheel odometry \cite{Junior_2022} and IMU measurements \cite{Xu2022}) using an extended Kalman filter or a factor graph for robust estimation. 
This approach enables efficient and precise sensor localization owing to the high-definition point cloud data provided by recent range sensors. However, these single hypothesis methods heavily rely on the continuity of the point cloud data. They easily fail to continue tracking the sensor pose when the point cloud data are interrupted. Even a momentary data drop (e.g., 1 s) can cause a critical estimation failure, and it is almost impossible to deal with a longer interruption (i.e., kidnapping) using iterative-scan-matching-based methods.

\subsection{Monte Carlo Localization}

MCL is one of the most popular approaches to the 2D localization problem. It represents and estimates a state distribution with a finite set of state samples. In particular, a particle filter is the most representative non-parametric Monte Carlo Bayes filter used to estimate the state distribution through particle importance weighting and resampling \cite{probrobo}. Because of its non-linear and non-Gaussian nature, it is robust to ambiguity in the observations and environment. With a sufficient number of samples, it can even enable pinpoint sensor localization without any prior knowledge (i.e., global localization). 

Despite its success in 2D localization, the MCL approach has not been commonly used in 3D scenarios because of the curse of dimensionality. Because the number of samples to fill a unit space grows exponentially with the dimensionality, a massive number of particles is needed, with a large computation cost, to retain the robust properties of the 2D MLC approaches in 3D scenarios. Most of the existing 3D MCL methods perform state estimation in 3 or 4 DoF (\cite{Saarinen_2013, Perez_Grau_2017}).

Recently, several studies have proposed 6-DoF MCL methods that smartly sample and evaluate particles. Akai et al. used a 3D distance field for accelerated likelihood evaluation and performed 6-DoF sensor pose estimation with a vehicle motion model \cite{Akai_2020}. Maggio et al., used a NeRF(Neural radiance field)-based map representation and VIO(visual inertial odometry)-based precise motion prediction \cite{Maggio_2023}. Deng et al. decoupled the rotation and translation components and estimated a 6-DoF object pose using a Rao-Blackwellized particle filter \cite{Deng_2021}. Maken et al. proposed a Stein particle filter that introduced SVGD to improve the sampling efficiency and demonstrated 4-DoF global localization and 6-DoF pose tracking on a 3D map \cite{Maken_2022}. Although these methods enabled 6-DoF state estimation with only 50-1,000 particles using smart sampling schemes, this number of samples made it difficult to represent a multi-modal distribution in a 6-DoF space. Thus, re-localization and ambiguity handling abilities were limited.

\subsection{GPU-Accelerated Localization}

Several studies have used GPU acceleration to achieve fast and robust localization. However, most of them used the GPU for only a portion of the system (e.g., nearest neighbor search \cite{Sun_2020} or occupancy gridmap generation \cite{Stepanas_2022}).

Our work was inspired by the work of Peng and Weikersdorfer, who proposed 2D localization with a histogram filter that used a belief tensor representing all the possible states \cite{Peng_2020}. The entire system was designed to leverage the GPU computation power, and they showed that a non-parametric filter accelerated with the GPU exhibited an excellent pinpoint global localization ability without any initial state assumption. Inspired by their work, we fully utilized GPU acceleration to achieve 6-DoF localization with a global re-localization ability.

\section{Methodology}

\subsection{Problem Setting}

Our objective is to estimate SE3 sensor pose ${\bm T}_t$ on 3D point cloud map $\mathcal{M} = \left\{ {\bm p}_k^M \in \mathbb{R}^3 \mid_{k = 1, \ldots, N^M} \right\}$ from sensor point cloud measurements $\mathcal{P}_t = \left\{ {\bm p}_k^S \in \mathbb{R}^3 \mid_{k = 1, \ldots, N^S} \right\}$.

As with a conventional particle filter, we estimate the state distribution using a set of samples (i.e., particles) $\mathcal{X}_t = \{ {\bm x}_t^i \mid_{i = 1, \ldots, N^P} \}$ and update the particle states through prediction and correction steps. In our case, each particle ${\bm x}_t^i$ represents a hypothesis of the sensor pose: ${\bm x}_t^i \coloneqq {\bm T}_t^i$. Inspired by \cite{Maken_2022}, we employ SVGD-based particle optimization in the correction step for efficient and robust state sampling. After the correction step, we explicitly estimate the posterior probability of each particle through a Bayesian filter over a neighbor particle graph.

\subsection{Prediction Step}

To update the particle states in the prediction step, we first perform GICP scan matching \cite{segal2009generalized} between consecutive scan point clouds $\mathcal{P}_{t-1}$ and $\mathcal{P}_t$ to obtain an estimate of the sensor motion $\Delta {\bm T}_t \sim {\bm T}_{t - 1}^{-1} {\bm T}_t$. We also estimate the covariance matrix $\Sigma^{\Delta {\bm T}} \in \mathbb{R}^{6 \times 6}$ based on the Hessian matrix of the scan matching optimization result.

Then, the states of particles are updated as follows:
\begin{align}
{\bm T}_t^i &= {\bm T}_{t-1}^i \Delta {\bm T}_t \exp \left( {\bm \delta}^i \right),
\end{align}
where ${\bm \delta}^i \sim \mathcal{N} (0, \Sigma^{\Delta {\bm T}})$ is a random noise in the tangent space of $\Delta {\bm T}_t$. 

\subsection{Correction Step}

{\bf Likelihood evaluation:} We employ the GICP distribution-to-distribution distance \cite{segal2009generalized} for the log likelihood function $\log p(\mathcal{P}_t | {\bm x}_t^i)$. In GICP, each point ${\bm p}_k^*$ is modeled as a Gaussian distribution $\mathcal{N}({\bm \mu}_k^*, {\bm \Sigma}_k^*)$ representing the local geometrical shape. For each scan point ${\bm p}_k^S$, we find the nearest map point ${\bm p}_k^M$ and compute the log likelihood as follows:
\begin{align}
\label{eq:likelihood}
& \log p(\mathcal{P}_t | {\bm x}_t^i) = - \sum_k {\bm e}_k^\top {\bm \Omega}_k {\bm e}_k, \\
& {\bm e}_k = {\bm \mu}_k^M - {\bm T}_t^i {\bm \mu}_k^S, \quad {\bm \Omega}_k = \left( {\bm \Sigma}^M_k + {\bm T}_t^i {\bm \Sigma}^S_k ({\bm T}_t^i)^\top \right)^{-1}.
\end{align}

For an efficient nearest map point search, we precompute a nearest neighbor field $m^\text{nnf}$ that voxelizes the map space at a specific resolution (e.g., 0.1 m for indoors, 0.2 m for outdoors) and stores the index of the nearest map point for each voxel. When computing the log likelihood, we look up $m^\text{nnf}$ to find the nearest map point of each scan point.

{\bf Particle state update:} We propose Gauss-Newton-based approximate SVGD particle optimization to update particle states with the likelihood function.

Because $\log p(\mathcal{P}_t|{\bm x}_t^i)$ is in the least squares form, as in Gauss-Newton optimization, we can obtain an optimal particle displacement vector ${\bm \psi}^i = {\bm H}^{-1} {\bm b}$ to maximize the log likelihood through quadratic approximation, where
\begin{align}
{\bm H} &= \sum_k {\bm J}_k^\top {\bm \Omega}_k {\bm J}_k, &{\bm b} &= \sum_k {\bm J}_k^\top {\bm \Omega}_k {\bm e}_k, & {\bm J_k} &= \frac{\partial {\bm e}_k} {\partial {\bm T}_t^i}.
\end{align}
Then, we update the states of particles based on modified SVGD:
\begin{align}
{\bm T}_{t + 1}^i &= {\bm T}_t^i  \exp \left( {\bm \phi} \left( {\bm T}_t^i, \mathcal{P}_t \right) \right), \\
\label{eq:phi}
{\bm \phi}({\bm T}_{t}^i, \mathcal{P}_t) &= \frac{ \sum_{ {\bm x}_t^j \in \widetilde{\mathcal{X}}_t^i } \left( k({\bm T}^i_t, {\bm T}^j_t) {\bm \psi}^j + \nabla_{{\bm T}_t^j} k({\bm T}_t^i, {\bm T}_t^j) \right) }{ \sum_{ {\bm x}_t^j \in \widetilde{\mathcal{X}}_t^i } k({\bm T}_t^i, {\bm T}_t^j) },
\end{align}
where $k$ is a positive definite kernel and $\widetilde{\mathcal{X}}_t^i$ is a set of neighbor particles, which includes ${\bm x}_t^i$ itself. Kernel $k$ measures the distance between particles and controls how particles affect each other. We use the following exponential kernel with diagonal weighting matrix ${\bm W}^K = \text{diag}( [\sigma_r, \sigma_r, \sigma_r, \sigma_t, \sigma_t, \sigma_t] )$ (e.g., $\sigma_r = 5.0$ $\text{rad}^{-1}$ and $\sigma_t = 2.5$ $\text{m}^{-1}$):
\begin{align}
k \left( {\bm T}_t^i, {\bm T}_t^j \right) &= \exp \left( -{\bm d}_{ij}^\top {\bm W}^K {\bm d}_{ij} \right), \\
{\bm d}_{ij} &= \log \left( \left({\bm T}_t^i \right)^{-1} {\bm T}_t^j \right).
\end{align}

Intuitively, the component $\sum k({\bm T}^i_t, {\bm T}_t^j) {\bm \psi}^j$ in Eq. \ref{eq:phi} is a weighted sum of the Gauss-Newton update vectors of neighboring particles that pushes particle ${\bm x}_t^i$ toward the modes of the likelihood function. $\nabla_{{\bm T}_t^j} k({\bm T}_t^i, {\bm T}_t^j)$ is the gradient of the kernel at the particle location that causes repulsive forces between particles and helps maintain the diversity of the particles in the state space. 

The original SVGD computes the update direction of a particle using all the other particles, resulting in a computational burden with a large number of particles. Considering that only neighbor particles have large kernel values and affect the update direction in SVGD, we only use up to $K$ neighbor particles $\widetilde{\mathcal{X}}_t^i$ (e.g., 20 neighbors) to compute the update vector in Eq. \ref{eq:phi}.

{\bf LSH-based iterative neighbor particle search:} Because we handle a massive amount of dynamically changing particles in the non-Euclidean SE3 space, conventional nearest neighbor methods (e.g., linear search and spatial-partitioning) cannot be applied to our problem.

To efficiently find neighbor particles, we use an iterative neighbor particle search using locality sensitive hashing (LSH) based on stable distributions \cite{Datar_2004}. Algorithm \ref{alg:neighbors} describes the proposed neighbor particle search algorithm. We first distribute particles into a hash table with $N^B$ buckets using the following hash function (Line 3 -- 5):
\begin{align}
  f^\text{LSH} ({\bm T}_t^i) &= \text{hash} \left( [ \lfloor {\bm \zeta}[i] \rfloor |_{i=1 \cdots 6} ] \right), \\
  {\bm \zeta} &= \alpha {\bm W}^K \log \left( \left({\bm T}^\text{LSH}\right) ^{-1} {\bm T}_t^i \right) + {\bm \delta}_i^\text{LSH},
\end{align}
where $\text{hash}$ is a function used to compute a hash value from a tuple of integers \cite{teschner2003optimized}, ${\bm T}^\text{LSH}$ is a random transformation that defines the tangent space to compute the hash value, ${\bm \delta}^\text{LSH} \sim \mathcal{N}( 0, \Sigma^\text{LSH} )$ is a random Gaussian noise, and $\alpha$ is a constant. 

\begin{algorithm}[tb]
\footnotesize
\caption{IterativeNeighborParticleSearch}
\label{alg:neighbors}
\begin{algorithmic}[1]
\State ${\bm T}^\text{LSH} \leftarrow \text{Random SE3 transformation}$
\State $\mathcal{B} \leftarrow \left[ \emptyset \mid_{i = 1, \ldots, N^B} \right] $  \Comment{LSH buckets}
\For{ ${\bm x}_t^i \in \mathcal{X}_t$ }                                     \Comment{Distribute loop}
  \State $h \leftarrow f^\text{LSH} ({\bm x}_t^i) \mod N^B$
  \State $\mathcal{B}[h] \leftarrow \mathcal{B}[h] \cup {\bm x}_t^i$    \Comment{Add particle to the bucket}
\EndFor
\For{ ${\bm x}_t^i \in \mathcal{X}_t$ }                                     \Comment{Gather loop}
  \State $h \leftarrow f^\text{LSH} ({\bm x}_t^i) \mod N^B$
  \For{ ${\bm x}_t^j \in \mathcal{B}_t[h]$ }
    \State $\tilde{\mathcal{X}}_t^i \leftarrow \tilde{\mathcal{X}}_t^i \cup {\bm x}_t^j$    \Comment{Add to the neighbor list}
    \If{ $| \tilde{\mathcal{X}}_t^i | > K$ }
      \State $\hat{\bm x} \leftarrow \argmin_{{\bm x}_t^k \in \tilde{\mathcal{X}}_t^i} k ( {\bm x}_t^i, {\bm x}_t^k )$  \Comment{Farthest particle}
      \State $\tilde{\mathcal{X}}_t^i \leftarrow \tilde{\mathcal{X}}_t^i \setminus \hat{\bm x}$  \Comment{Remove $\hat{\bm x}$ from the neighbor list}
    \EndIf
  \EndFor
\EndFor
\end{algorithmic}
\end{algorithm}

\begin{figure}[tb]
  \centering
  \includegraphics[width=0.4\linewidth]{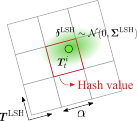}
  \caption{SE3 locality sensitive hashing based on a stable distribution.}
  \label{fig:lsh}
\end{figure}

As shown in Fig. \ref{fig:lsh}, ${\bm T}^\text{LSH}$ defines the grid to discretize transformation ${\bm T}_t^i$, and noise ${\bm \delta}^\text{LSH}$ causes a fluctuation in the position of ${\bm T}_t^i$ in the grid space. Because particles with a small displacement tend to fall in the same cell of the grid, $f^\text{LSH}$ assigns the same hash value to close particles with a high probability. By using $f^\text{LSH}$, we store all the particles in the hash table and then find particles that fall in a same bucket as neighbor particles. In this way, we can efficiently find neighbor particle candidates for all the particles in time linear to the number of particles (Line 6 -- 9).

However, the neighbor particles found will contain false positives and negatives because $f^\text{LSH}$ is a probabilistic function. We thus assume that the neighborhood relationship of particles does not significantly change in a short time interval, and we iteratively update the neighbor list $\widetilde{\mathcal{X}}_t^i$ of each particle over time by keeping only the $K$ closest particles (Line 9 -- 12).

\subsection{Posterior Propagation on Neighbor Particle Graph}

Similar to \cite{Maken_2022}, the SVGD-based correction step does not perform resampling. Because the proposed algorithm keeps all the particles alive, it does not suffer from the sample impoverishment problem \cite{Arulampalam_2002}. Meanwhile, a massive number of particles is obtained with an extremely non-linear distribution with many modes, and explicitly estimating the posterior distribution and representative state of the estimation result is not straight forward. We thus propose a method to explicitly estimate the posterior probability of each particle by propagating probabilities over a neighbor particle graph that is a by-product of the iterative neighbor particle search.

Given prior probability $p({\bm x}_t^i)$ of particle ${\bm x}_t^i$, we first obtain initial posterior probability $p({\bm x}_t^i | \mathcal{P}_t) \propto p({\bm x}_t^i) p(\mathcal{P}_t | {\bm x}_t^i)$ with likelihood $p(\mathcal{P}_t | {\bm x}_t^i)$ in Eq. \ref{eq:likelihood}. We then compute the weighted average of the neighbor particles:
\begin{align}
\label{eq:posterior}
p' ({\bm x}_t^i | \mathcal{P}_t) &= \frac{ \sum_{{\bm x}_t^j \in \widetilde{\mathcal{X}}_t^i} k({\bm T}_t^i, {\bm T}_t^j) p ({\bm x}_t^j | \mathcal{P}_t) }{ \sum_{{\bm x}_t^j \in \widetilde{\mathcal{X}}_t^i} k({\bm T}_t^i, {\bm T}_t^j) }.
\end{align}
We iteratively apply Eq. \ref{eq:posterior} several times (e.g., 10 times). This process can be interpreted as locally distributing and smoothing the posterior probabilities over the neighbor particle graph under the random walk assumption. The proposed Bayesian filtering approach can also be interpreted as a histogram filter with sparse and dynamic state bins updated with SVGD. After the posterior probability update, we obtain the state of the particle with the highest posterior probability as the representative state.

\section{Experiment}

\subsection{Indoor Experiment}

{\bf Experimental setting:} To demonstrate the robust initialization and re-localization ability of the proposed method, we conducted experiments in an indoor environment with repeated and symmetric structures. We used an Azure Kinect to acquire point clouds at 10 Hz. We recorded two sequences (Easy01 and Easy02) while walking in a corridor without aggressive motion and data interruption, and two other sequences (Kidnap01 and Kidnap02) with thee long data interruptions (10-20 s) in each by completely blocking the sensor view. Furthermore, the sensor was moved through room after room during the data interruptions, and we believe that none of the existing methods could deal with these severe kidnapping situations. The sequences (Easy01, Easy02, Kidnap01, Kidnap02) had durations of 139, 136, 147, and 109 s, respectively \footnote{The dataset is available at : \url{https://staff.aist.go.jp/k.koide/projects/icra2024_mp/}}.

To run the proposed algorithm, we initialized $1024^2$ particles with a uniform distribution covering the entire map (50 m $\times$ 35 m $\times$ 4 m and full SO3 rotation). All the particles were evaluated and updated in real-time on a single GPU (NVIDIA A100).

As a baseline, we ran two localization algorithms based on iterative scan matching, FAST\_LIO\_LOCALIZATION \footnote{\url{https://github.com/HViktorTsoi/FAST_LIO_LOCALIZATION}} and hdl\_localization \cite{Koide_2019}. FAST\_LIO\_LOCALIZATION uses FAST\_LIO2, which is a tightly coupled LiDAR-IMU odometry estimation method \cite{Xu2022}, to estimate the sensor ego-motion and periodically performs scan-to-map registration to correct estimation drift. For comparison, we also ran FAST\_LIO2 \cite{Xu2022} without map-based pose correction. hdl\_localization \cite{Koide_2019} performs NDT-based scan-to-map registration and unscented Kalman filter-based IMU fusion. Note that we used IMU data and initial poses only for these existing methods, whereas the proposed method performed localization using only point cloud data without initial guess.

To obtain reference sensor trajectories, we manually aligned scan point clouds with the map point cloud and performed batch optimization of the scan-to-map registration errors and IMU motion errors. We evaluated the estimated trajectories using the absolute trajectory error (ATE) metric \cite{Zhang_2018} using the {\it evo} toolkit \footnote{\url{https://github.com/MichaelGrupp/evo}}. Because we initialized the proposed method with a uniform distribution (no initial guess), we excluded the beginning of each sequence until the posterior probability distribution converged to a single pose hypothesis (10-30 s).

\begin{figure}[tb]
  \centering
  \begin{minipage}[tb]{0.32\linewidth}
  \centering
  \includegraphics[width=\linewidth]{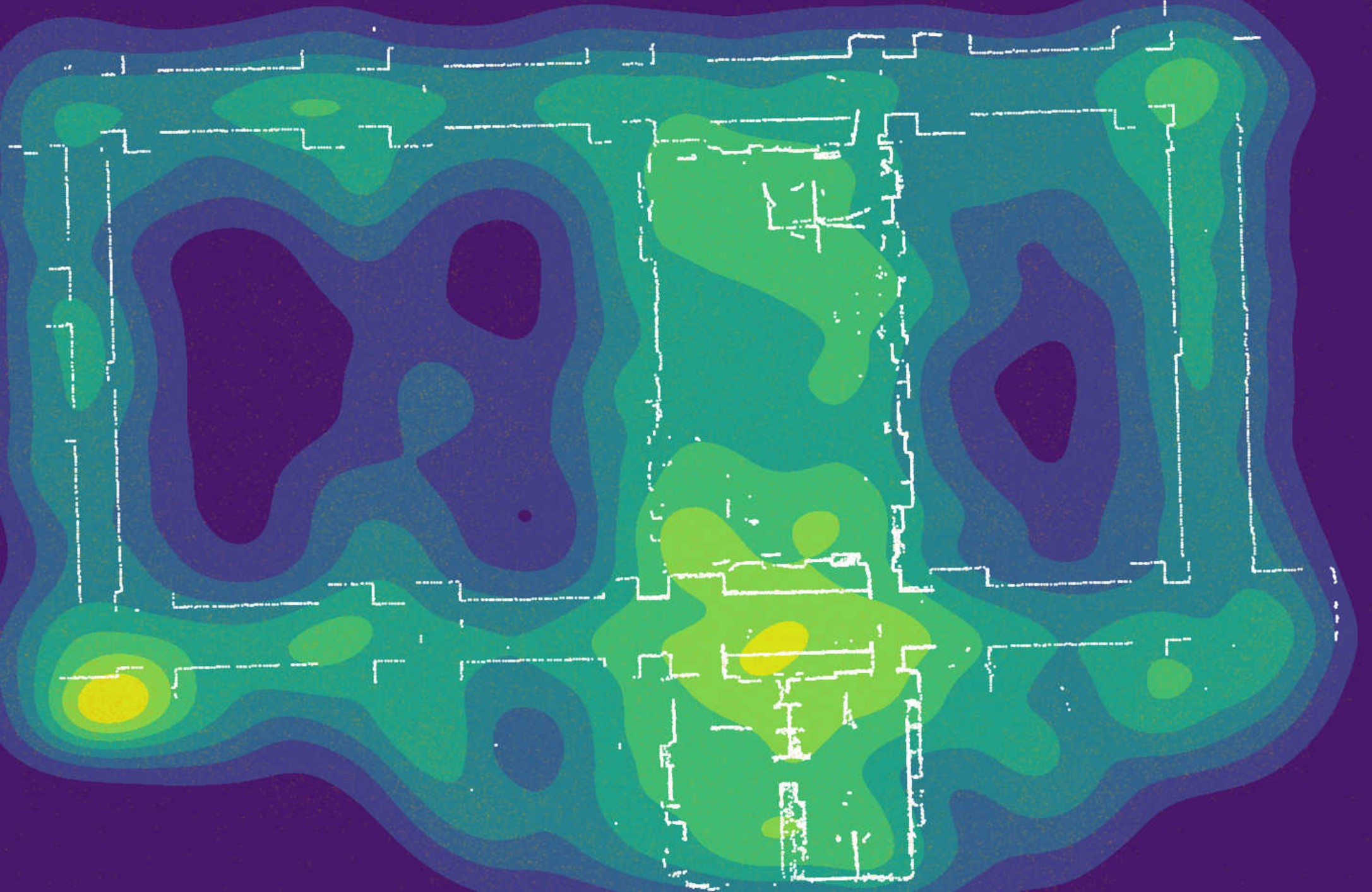}
  \subcaption{}
  \end{minipage}
  \begin{minipage}[tb]{0.32\linewidth}
  \centering
  \includegraphics[width=\linewidth]{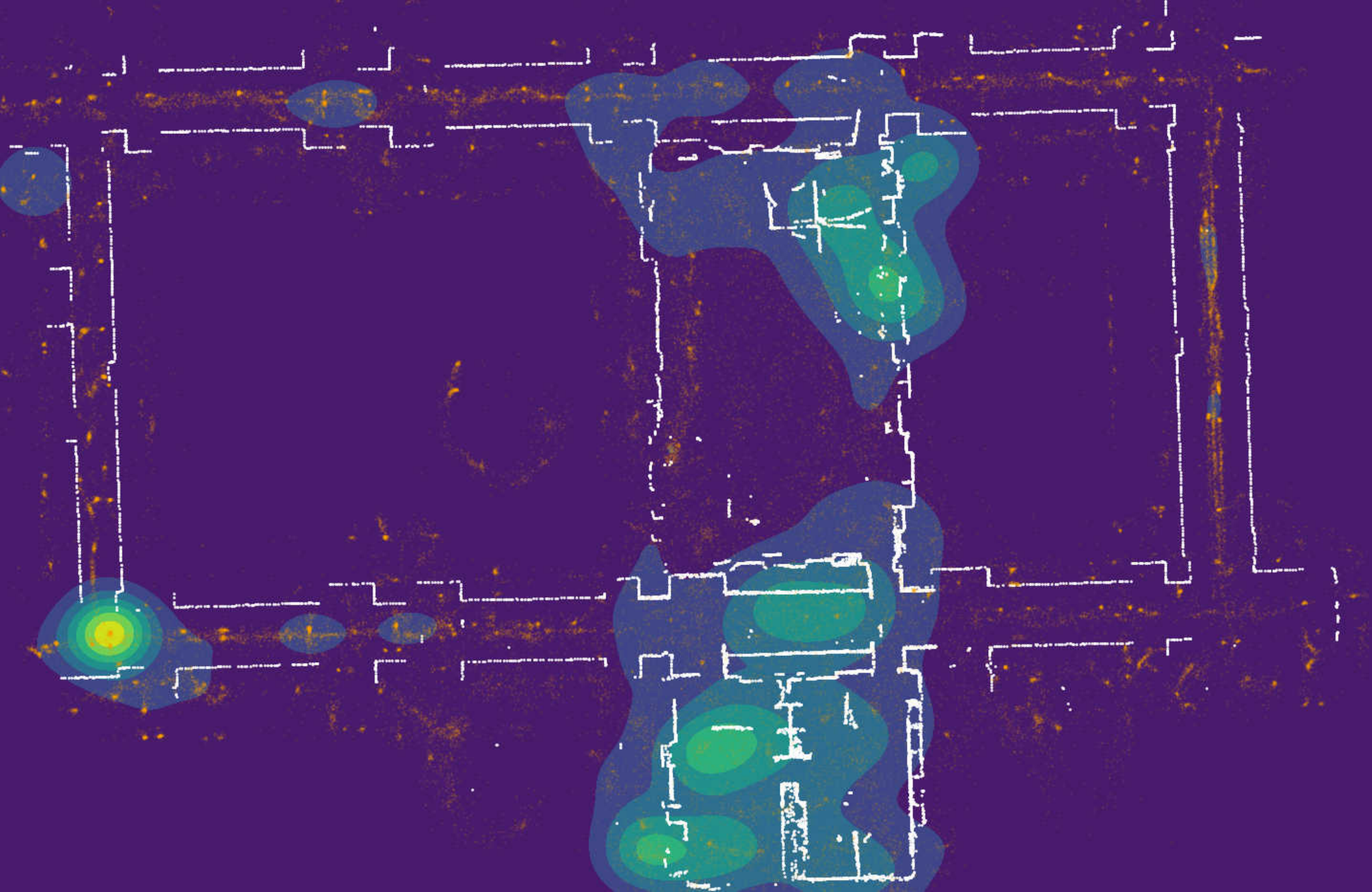}
  \subcaption{}
  \end{minipage}
  \begin{minipage}[tb]{0.32\linewidth}
  \centering
  \includegraphics[width=\linewidth]{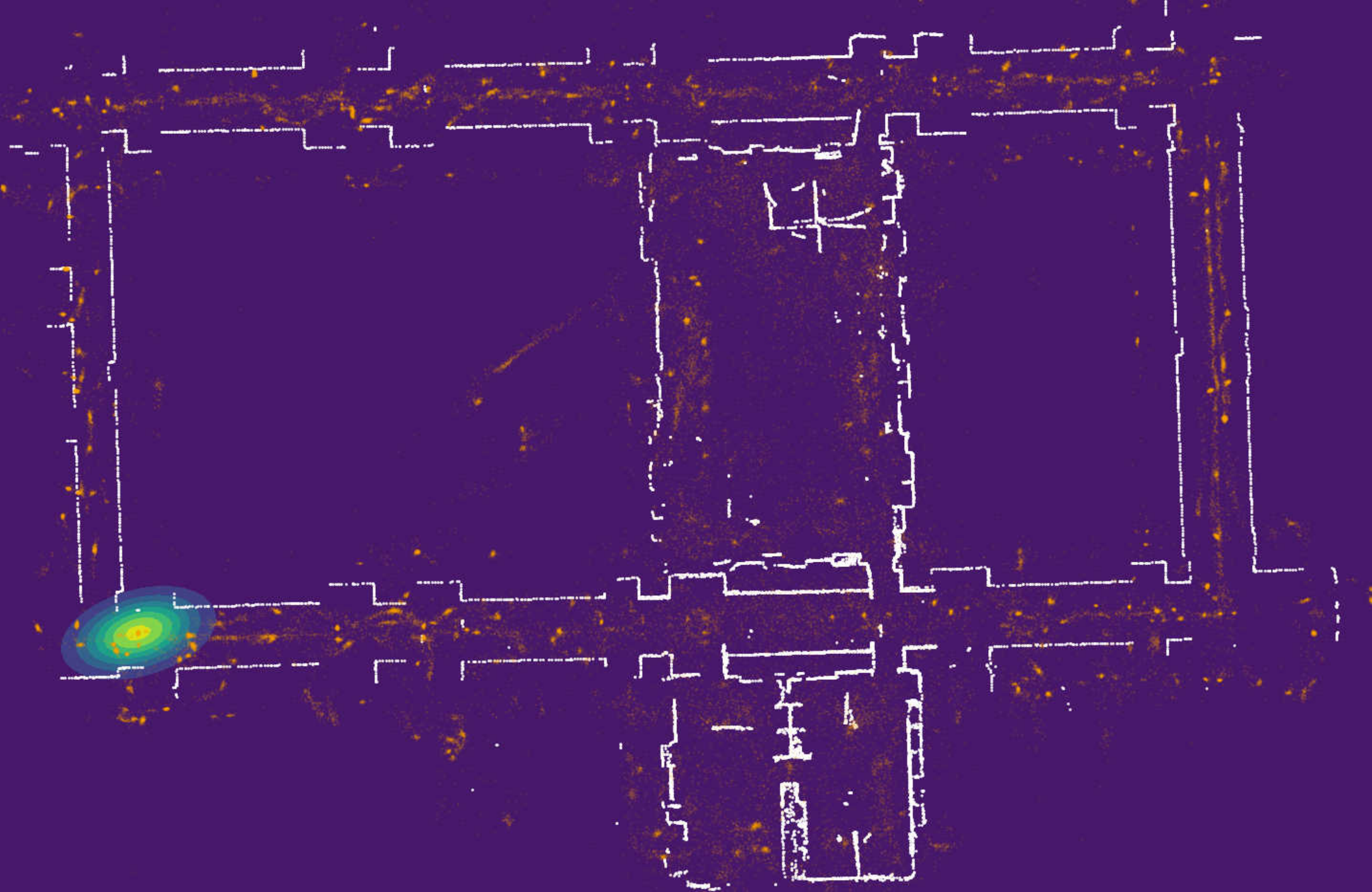}
  \subcaption{}
  \end{minipage}

  \begin{minipage}[tb]{0.32\linewidth}
  \centering
  \includegraphics[width=\linewidth]{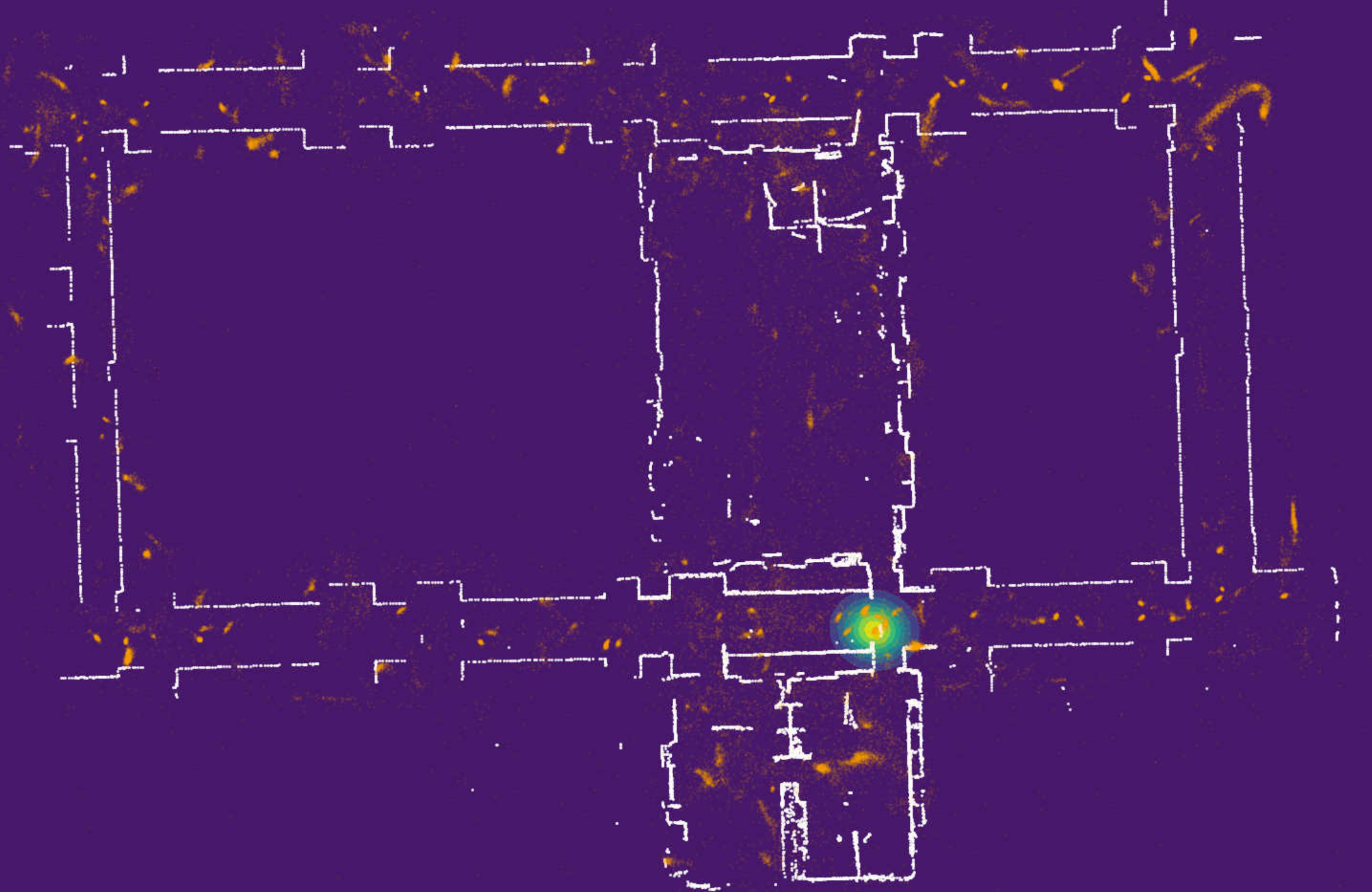}
  \subcaption{}
  \end{minipage}
  \begin{minipage}[tb]{0.32\linewidth}
  \centering
  \includegraphics[width=\linewidth]{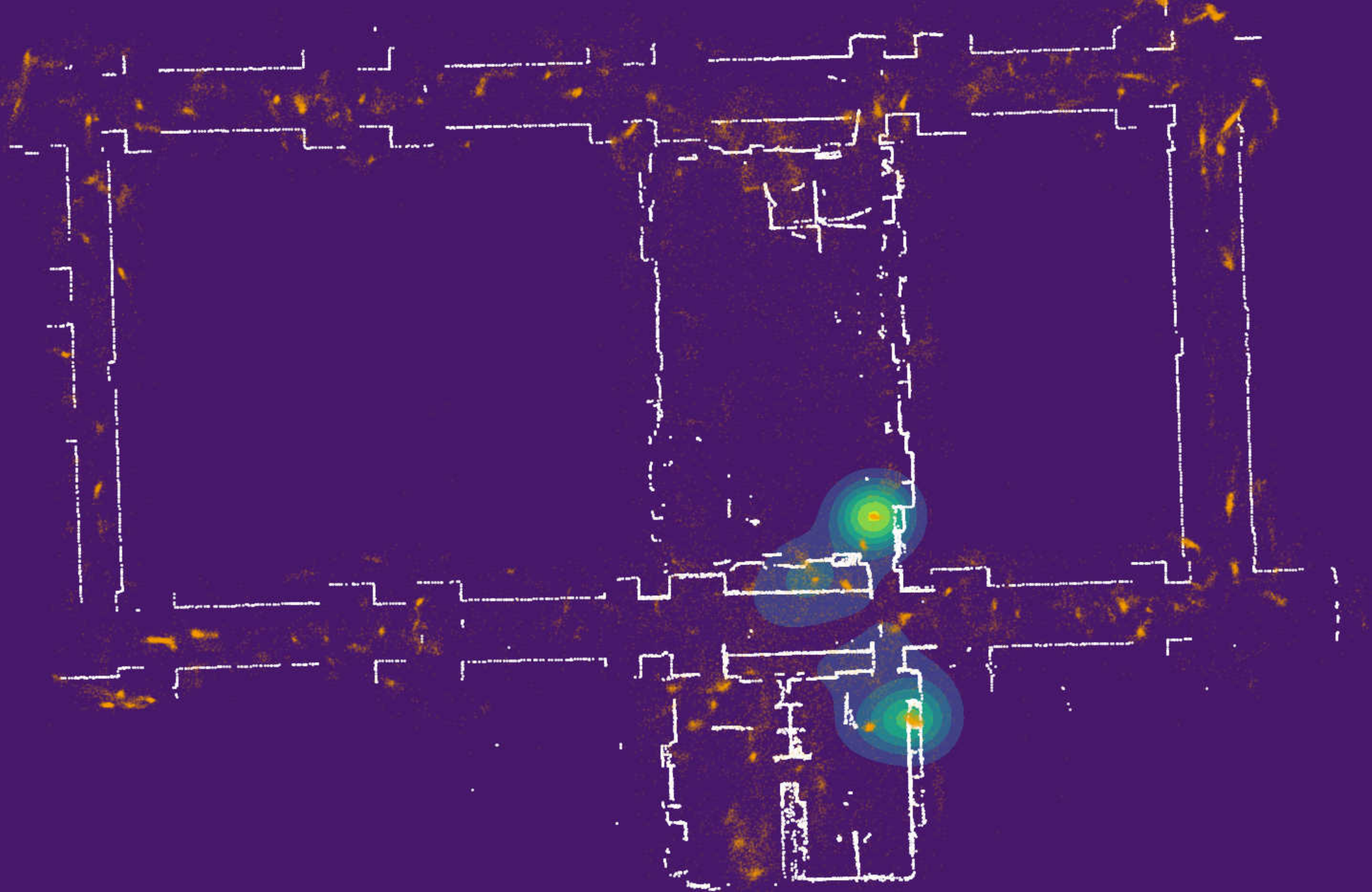}
  \subcaption{}
  \end{minipage}
  \begin{minipage}[tb]{0.32\linewidth}
  \centering
  \includegraphics[width=\linewidth]{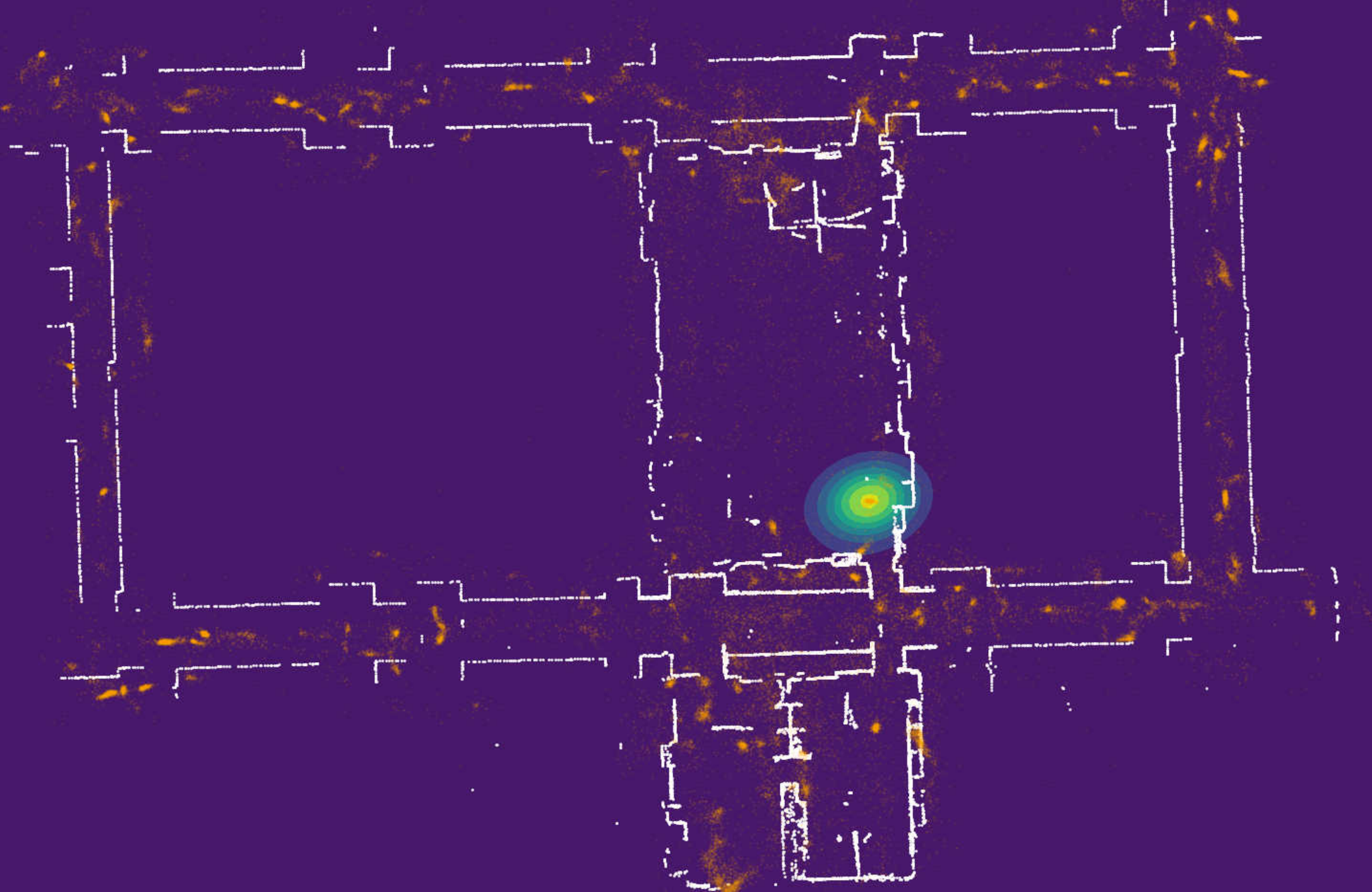}
  \subcaption{}
  \end{minipage}

  \caption{Estimation result for Easy01 sequence. Particles were initialized with a uniform distribution without any prior information (a). The positional ambiguity was quickly resolved as the sensor moved (b, c). Although there were still two major hypotheses for the sensor orientation (upright and flipped) (d, e), it was resolved when the sensor entered a room (f).}
  \label{fig:indoor_posterior_01}
\end{figure}

\begin{figure}[tb]
  \centering
  \begin{minipage}[tb]{0.32\linewidth}
  \centering
  \includegraphics[width=\linewidth]{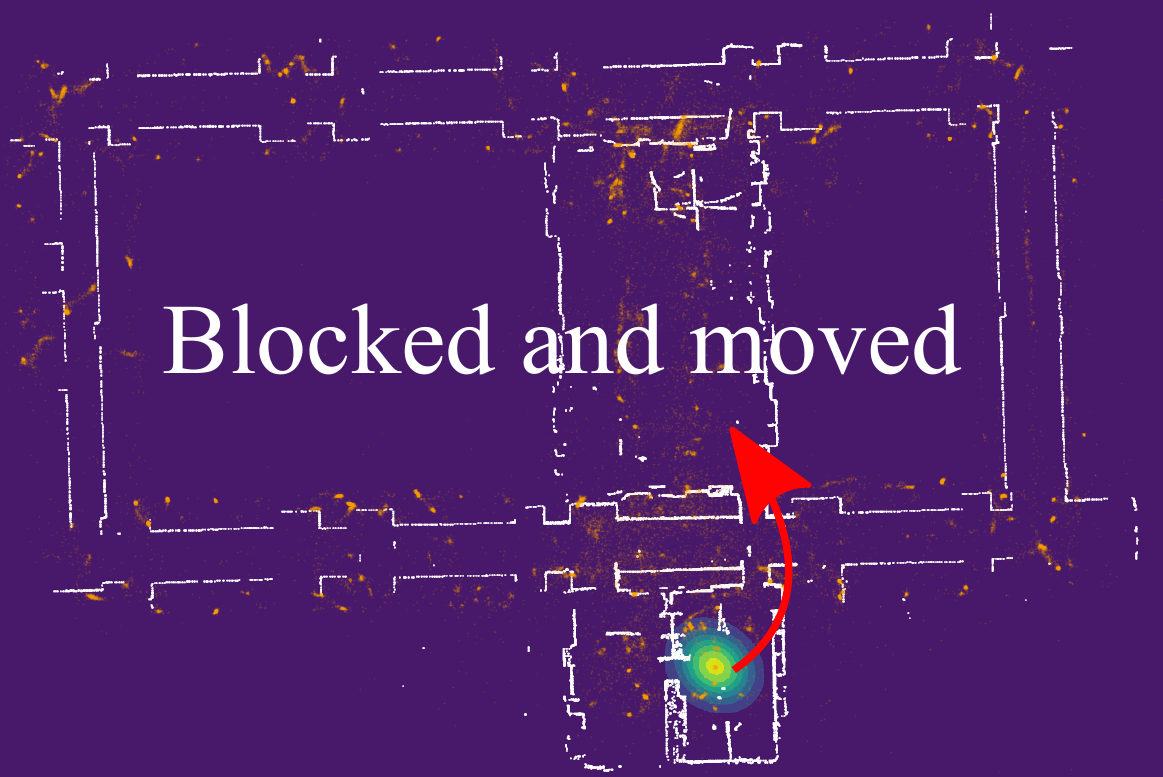}
  \subcaption{}
  \end{minipage}
  \begin{minipage}[tb]{0.32\linewidth}
  \centering
  \includegraphics[width=\linewidth]{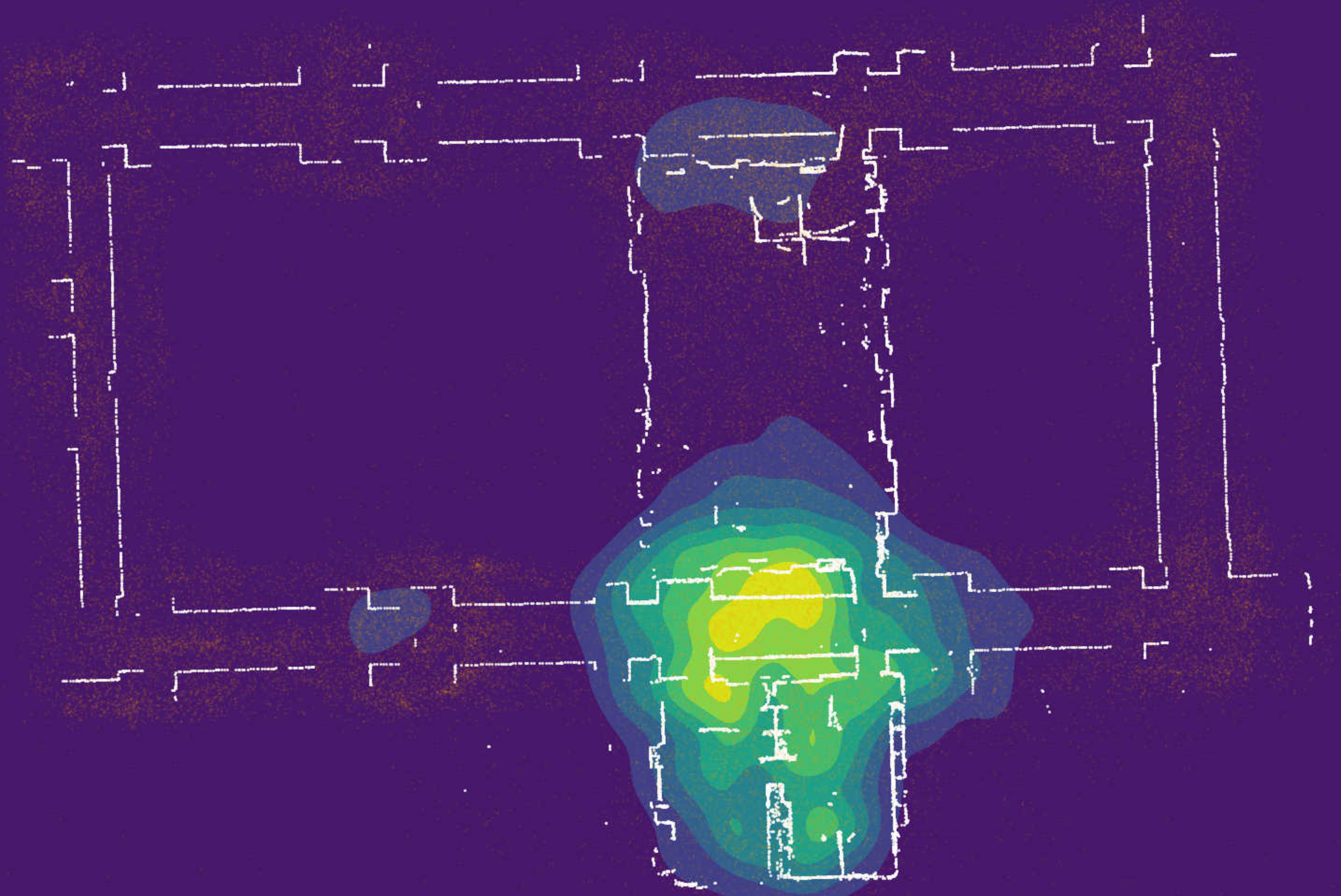}
  \subcaption{}
  \end{minipage}
  \begin{minipage}[tb]{0.32\linewidth}
  \centering
  \includegraphics[width=\linewidth]{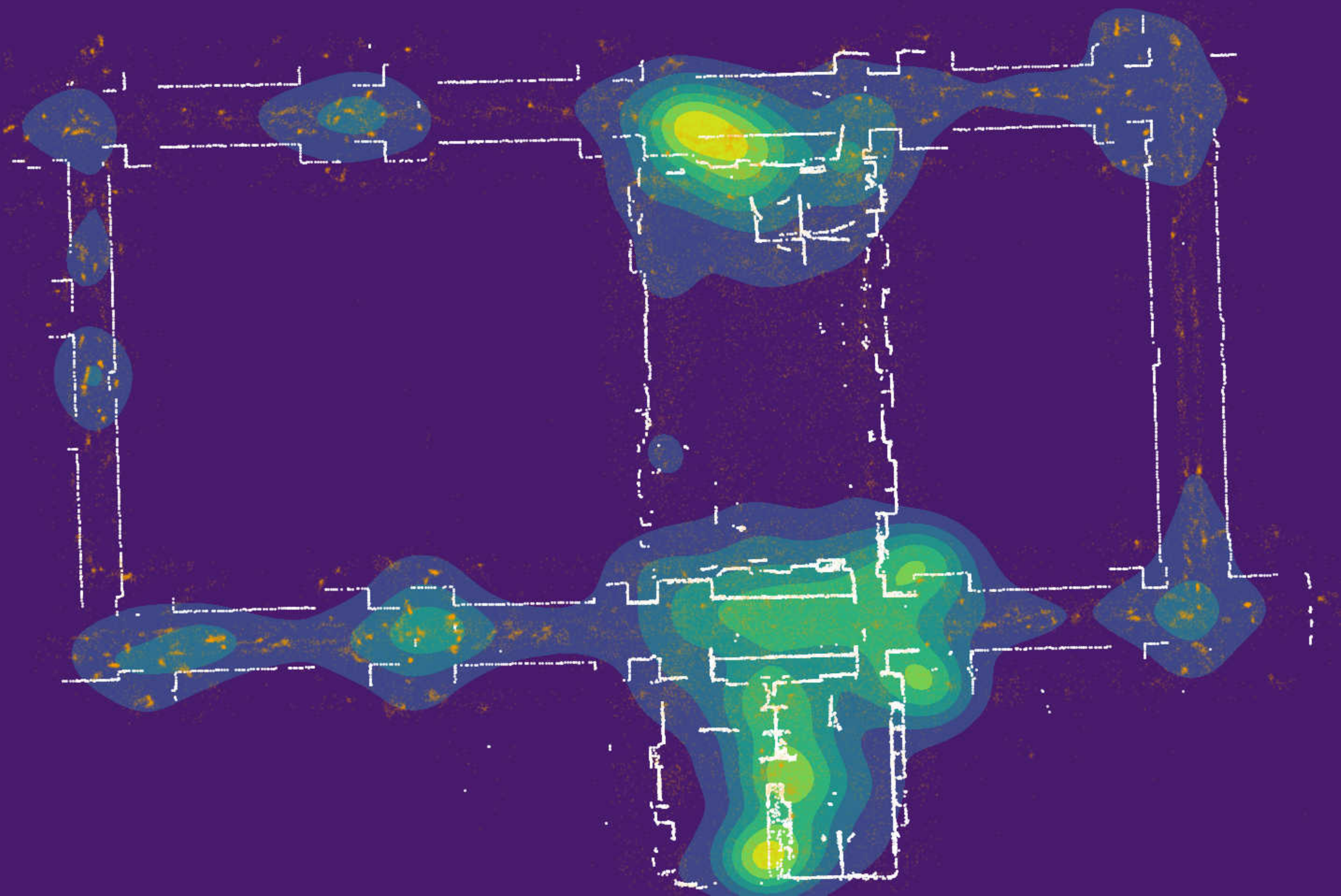}
  \subcaption{}
  \end{minipage}

  \begin{minipage}[tb]{0.32\linewidth}
  \centering
  \includegraphics[width=\linewidth]{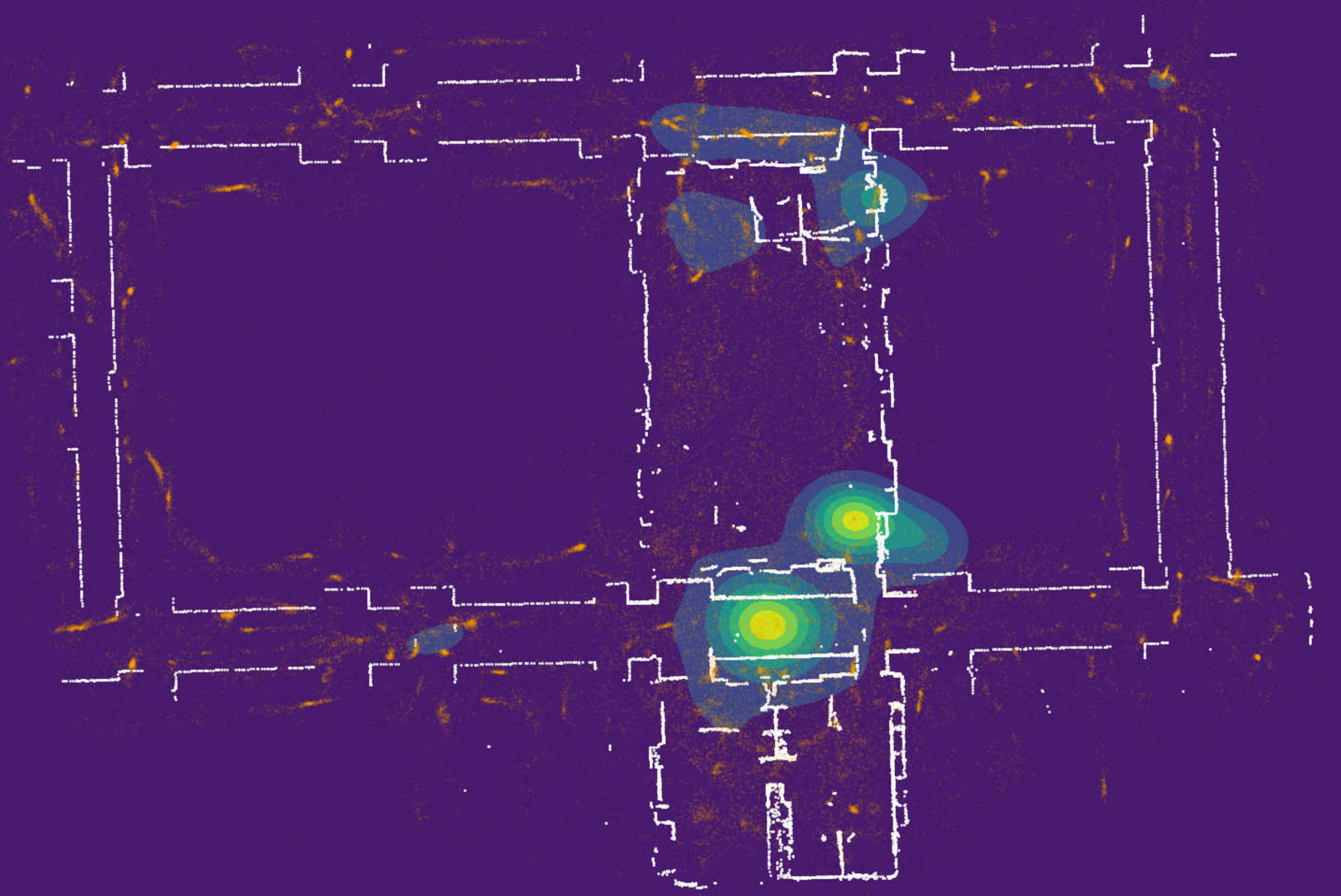}
  \subcaption{}
  \end{minipage}
  \begin{minipage}[tb]{0.32\linewidth}
  \centering
  \includegraphics[width=\linewidth]{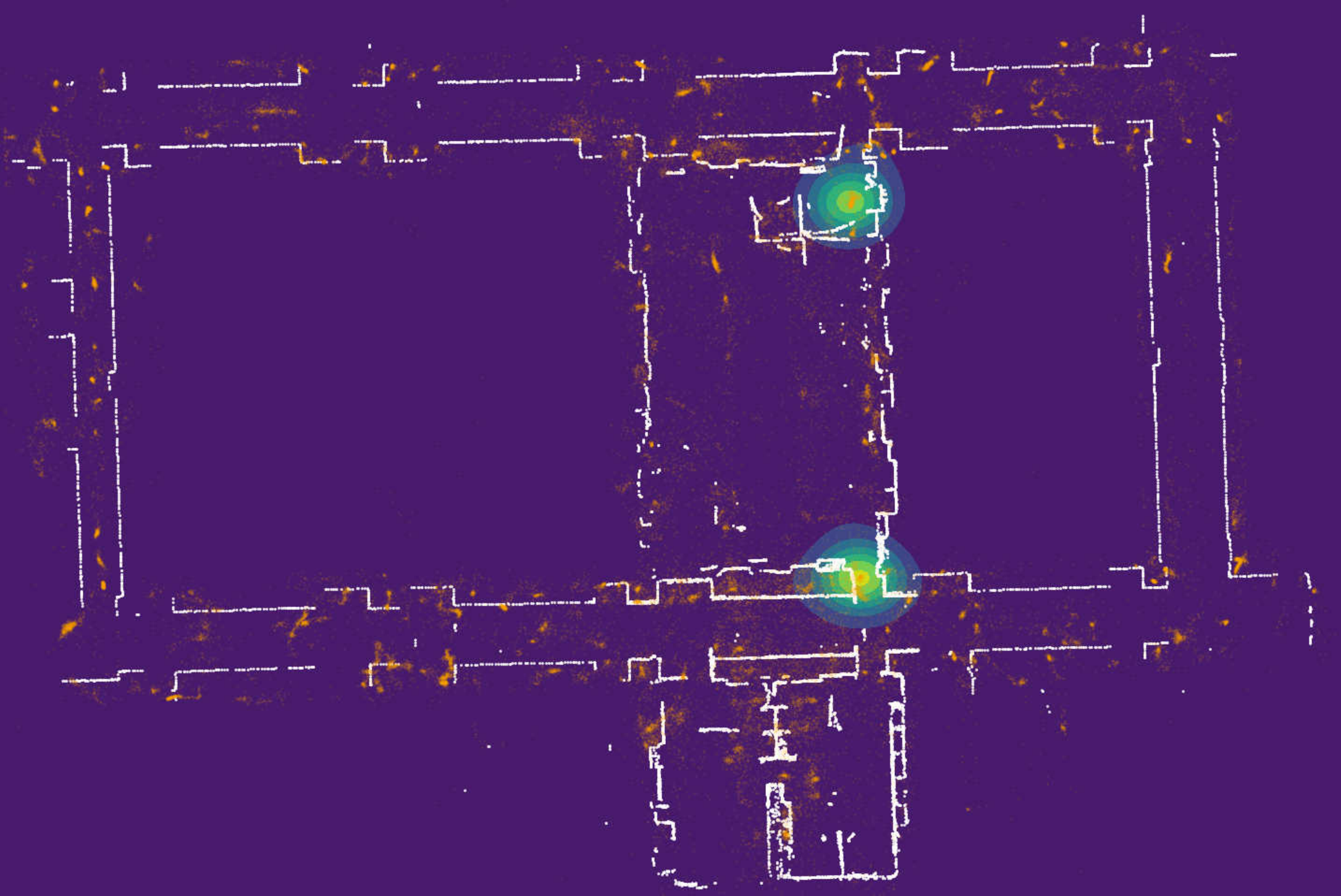}
  \subcaption{}
  \end{minipage}
  \begin{minipage}[tb]{0.32\linewidth}
  \centering
  \includegraphics[width=\linewidth]{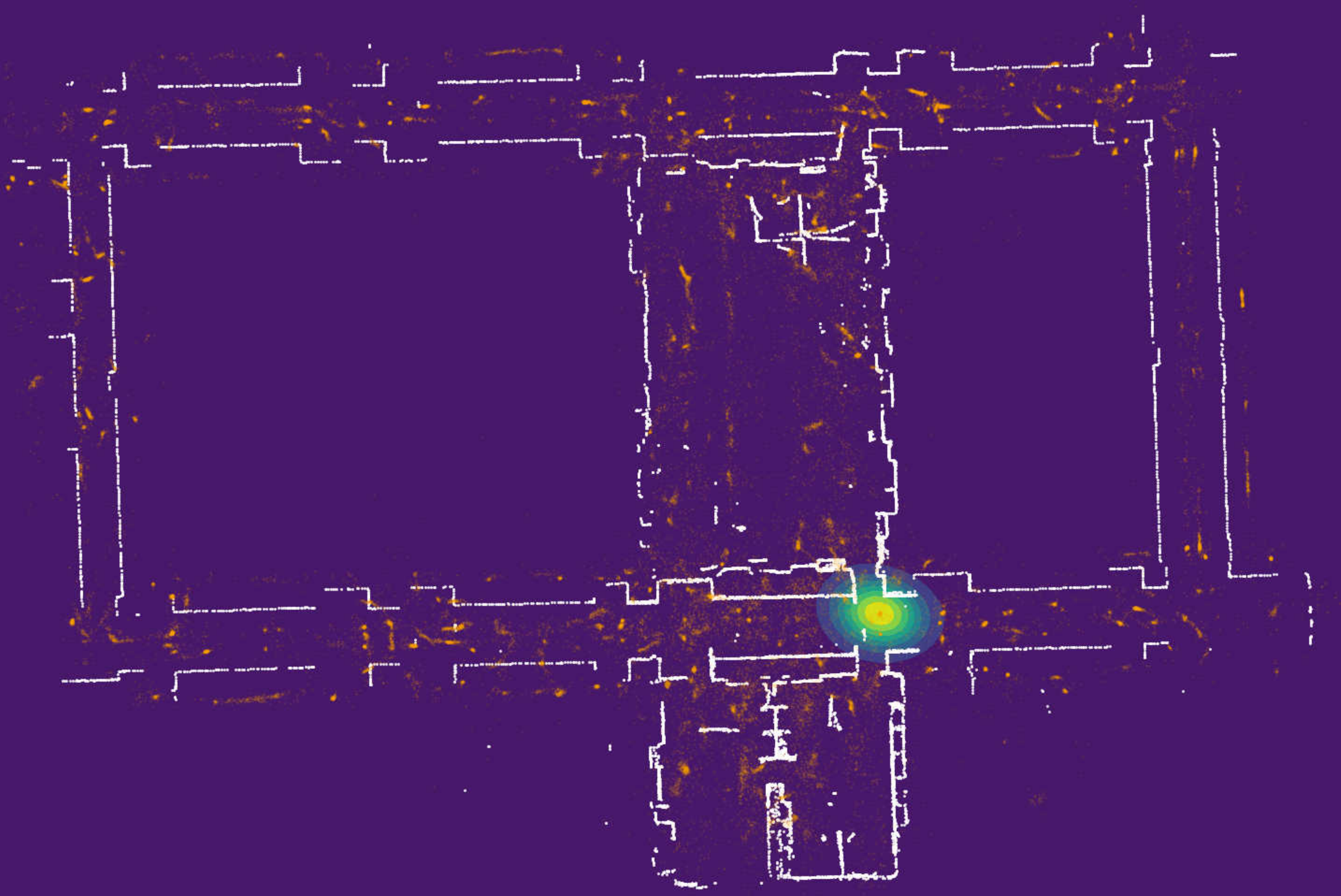}
  \subcaption{}
  \end{minipage}

  \caption{Kidnapping experiment result (Kidnap02). The sensor view was completely occluded, and the uncertainty grew because of the lack of observations (a, b). When the occlusion was removed and the sensor began to see the world, the posterior probability distribution quickly converged to a few positions (c, d, e). Eventually, the posterior distribution successfully converged to the correct position (f).}
  \label{fig:indoor_posterior_07}
\end{figure}

{\bf Initialization:} Fig. \ref{fig:indoor_posterior_01} shows how the posterior probability distribution of the proposed method converged from a uniform distribution in the Easy01 sequence. We applied weighted kernel density estimation to the 2D particle positions for visualization. Although the particles were uniformly initialized without any prior information (Fig. \ref{fig:indoor_posterior_01} (a)), the posterior probability quickly converged around the correct position as the sensor moved (Fig. \ref{fig:indoor_posterior_01} (b)(c)). Although there was still orientation ambiguity (upright and flipped) due to the symmetric environment structure (Fig. \ref{fig:indoor_posterior_01} (d)), when the sensor made a turn and entered a room (Fig. \ref{fig:indoor_posterior_01} (e)), the ambiguity was eventually resolved (Fig. \ref{fig:indoor_posterior_01} (f)).

{\bf Dealing with kidnapping:} Fig. \ref{fig:indoor_posterior_07} shows how the proposed method recovered from kidnapping in the Kidnap02 sequence. During the data interruption, the particles were spread around, and the posterior distribution diverged (Fig. \ref{fig:indoor_posterior_07} (b)). Meanwhile, the sensor was moved to another room while the sensor view was kept completely blocked. Once the point cloud data became available, the particles spread over the map (Fig. \ref{fig:indoor_posterior_07} (c)) and then quickly converged to a few possible locations with structures similar to those in the observed scan points (Fig. \ref{fig:indoor_posterior_07} (d) (e)). Eventually, it converged around the correct pose (Fig. \ref{fig:indoor_posterior_07} (f)) and recovered from the kidnapping. In these experiments, the proposed method successfully recovered the tracking of the sensor pose after every kidnapping.

As expected, both FAST\_LIO\_LOCALIZATION and hdl\_localization failed to continue tracking the sensor pose in the Kidnap01 and Kidnap02 sequences because of the severe data interruptions, and their estimations eventually became corrupted. These results showed the superior re-localization ability of the proposed method. The non-Gaussian state estimation combined with a massive amount of particles made it possible to search for candidate sensor locations over the entire map, and the SVGD-based particle update made the particles quickly converge to the correct location.

\begin{figure}[tb]
  \centering
  \includegraphics[width=0.8\linewidth]{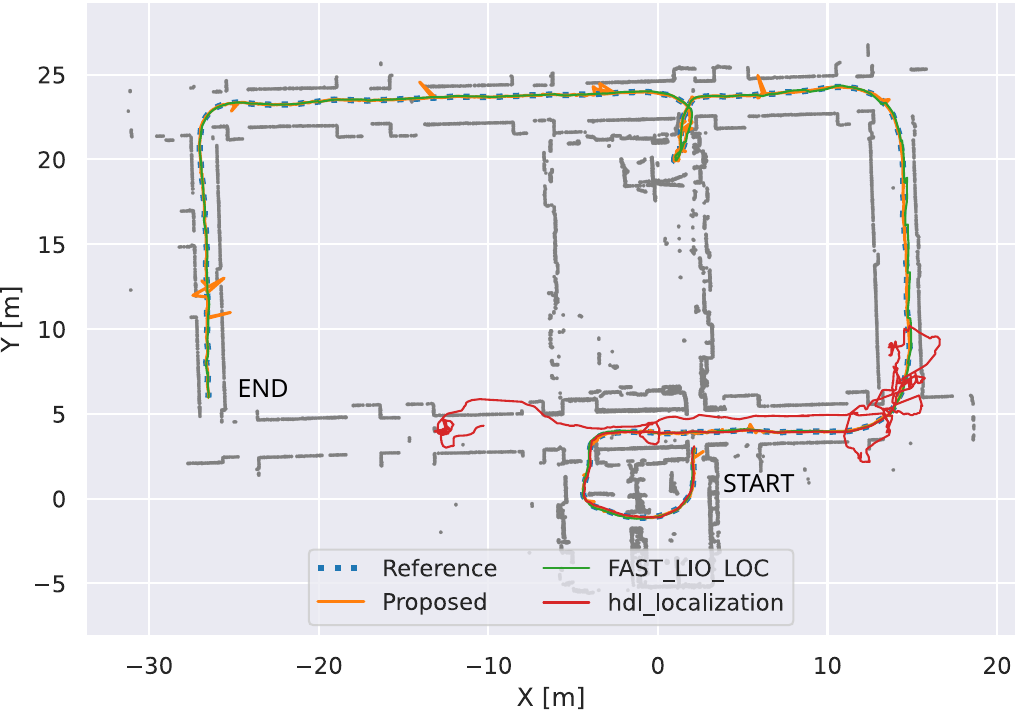}
  \caption{Estimated trajectories for the Easy02 sequence.}
  \label{fig:indoor_traj_02}
\end{figure}

{\bf Estimation accuracy:} Fig. \ref{fig:indoor_traj_02} shows the estimated trajectories for the Easy02 sequence. Once the posterior distribution converged around the correct pose, the proposed method robustly kept tracking the sensor pose until the end of the sequence. Although the estimation result showed momentary pose jitters because we simply chose the particle with the highest posterior probability as the representative state, these jitters could be filtered out using simple trajectory smoothing.

%
%
%
%

\newcommand{\cmark}{\ding{51}}%
\newcommand{\xmark}{\ding{55}}%
\setlength{\tabcolsep}{5pt}

\begin{table}[tb]
  \caption{Absolute trajectory errors for indoor sequences}
  \label{tab:results}
  \centering
  \scriptsize
  \begin{tabular}{l|cccc}
  \toprule
                    & \multicolumn{4}{c}{ATE [m]} \\
  Method            & Easy01          & Easy02          & Kidnap01        & Kidnap02 \\
  \midrule
  FAST\_LIO (odom)  & 1.86 $\pm$ 0.85 & 6.16 $\pm$ 3.02 & \xmark          & \xmark   \\
  FAST\_LIO\_LOC    & 0.07 $\pm$ 0.05 & 0.14 $\pm$ 0.10 & \xmark          & \xmark   \\
  hdl\_localization & 0.14 $\pm$ 0.10 & 16.8 $\pm$ 10.3 & \xmark          & \xmark   \\
  Proposed          & 0.25 $\pm$ 0.24 & 0.13 $\pm$ 0.12 & 5.64 $\pm$ 4.86 & 5.94 $\pm$ 5.00 \\
  \bottomrule
  \end{tabular}

  \xmark indicates that the estimation became corrupted.
\end{table}

Table \ref{tab:results} summarizes the ATE results for the evaluated methods. We can see that although FAST\_LIO without map-based correction showed large errors (1.86 m and 6.16 m) for Easy01 and Easy02, the errors were significantly reduced to 0.07 and 0.14 m with map-based correction, respectively. While hdl\_localization showed good accuracy for Easy01 (0.14 m), it showed a worse result for Easy02 (16.8 m) as a result of the tracking failure caused by the feature-less environment. Although the proposed method showed a slightly worse estimation result for the Easy01 sequence due to pose jitters, these results were generally comparable to those of the existing methods (0.25 m and 0.13 m).

\begin{figure}[tb]
  \centering
  \includegraphics[width=0.8\linewidth]{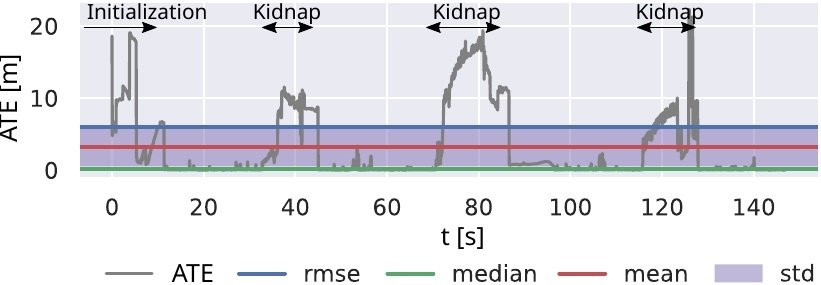}
  \caption{ATE of the proposed method for Kidnap02 sequence.}
  \label{fig:ate_kidnap}
\end{figure}

While the existing methods became corrupted in the Kidnap01 and Kidnap02 sequences, the proposed method successfully kept tracking the sensor pose through these sequences. Fig. \ref{fig:ate_kidnap} shows a plot of the ATE results for the proposed method during the Kidnap02 sequence. We can see that, at the beginning, the proposed method quickly converged to the correct location, and the ATE decreased. Although the ATE increased during kidnapping, once the point cloud data became available, the estimation converged to the correct sensor locations and showed small ATE values.

{\bf Processing time:}  Table \ref{tab:indoor_time} summarizes the processing time of each step in the proposed method. The entire process was highly parallelized and performed on a single GPU in real-time, and it took approximately 91 ms for each scan frame on average.


\begin{table}[tb]
  \caption{Processing time for indoor environments}
  \label{tab:indoor_time}
  \centering
  \scriptsize
  \begin{tabular}{l|l}
  \toprule
  Process                       & Time [ms] \\
  \midrule
  Neighbor list update          & 26.67 $\pm$ 0.35 \\
  Likelihood evaluation         & 55.17 $\pm$ 3.84 \\
  Particle state update         & 1.59  $\pm$ 0.03 \\
  Posterior probability update  & 7.30  $\pm$ 1.02 \\
  \midrule
  Total                         & 90.8  $\pm$ 4.20 \\
  \bottomrule
  \end{tabular}
\end{table}

\subsection{Outdoor Experiment}

\begin{figure}[tb]
  \centering
  \includegraphics[width=1.0\linewidth]{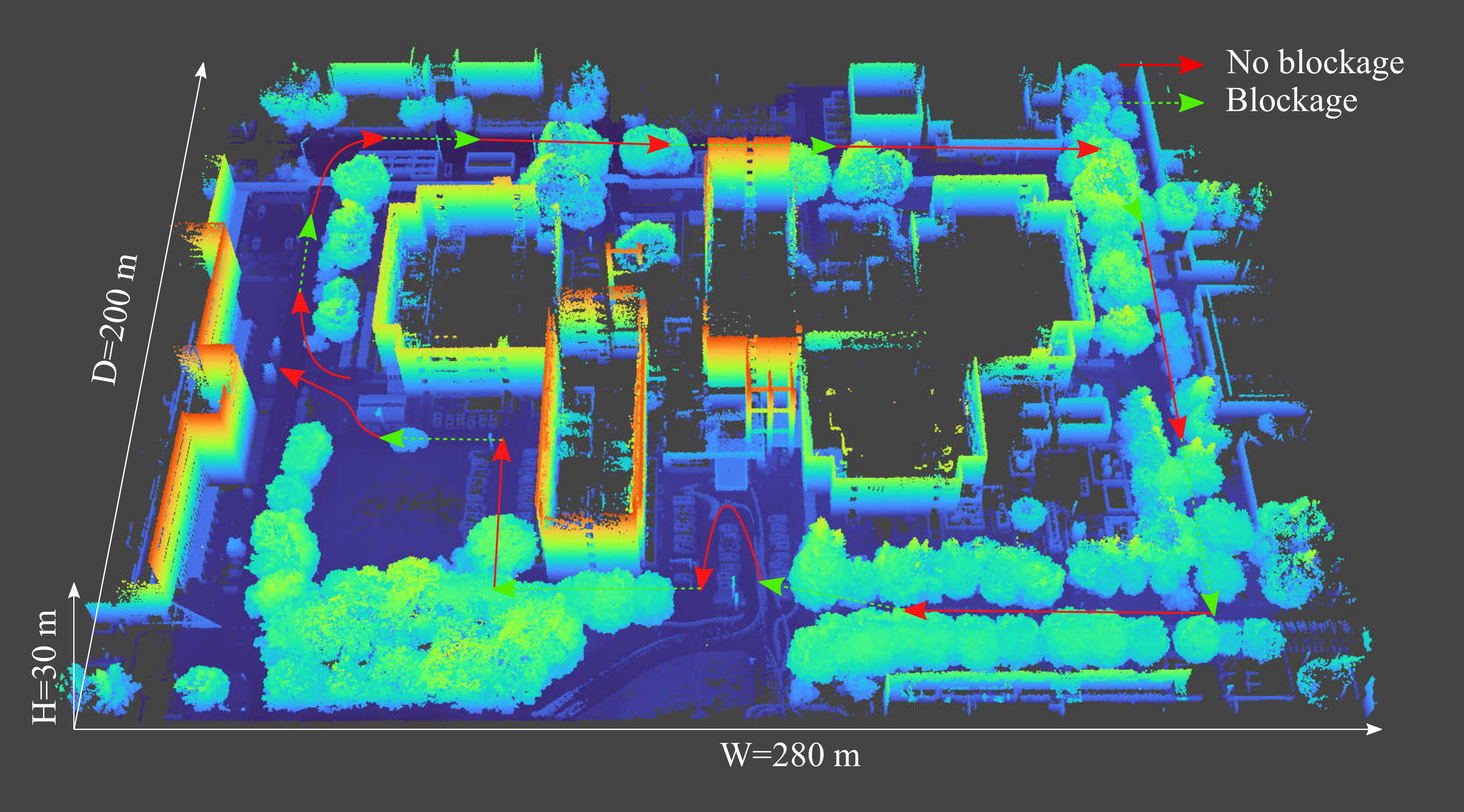}
  \caption{Outdoor experimental environment. During the experiment, complete blockages of the LiDAR (Livox MID-360) made the point cloud feed unavailable for 10-30 s.}
  \label{fig:outdoor}
\end{figure}

\begin{figure}[tb]
  \centering
  \begin{minipage}[tb]{0.45\linewidth}
  \centering
  \includegraphics[width=\linewidth]{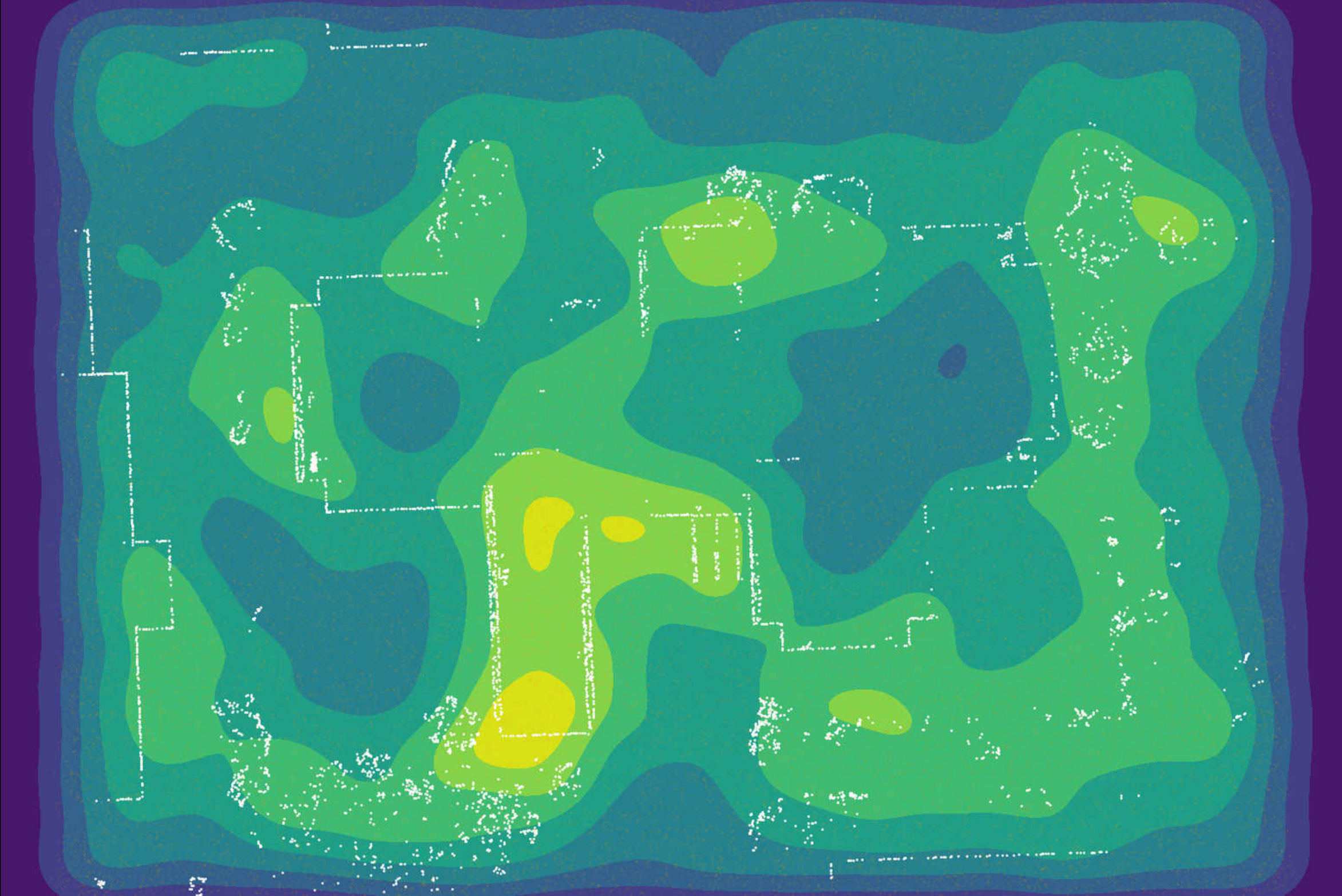}
  \subcaption{}
  \end{minipage}
  \begin{minipage}[tb]{0.45\linewidth}
  \centering
  \includegraphics[width=\linewidth]{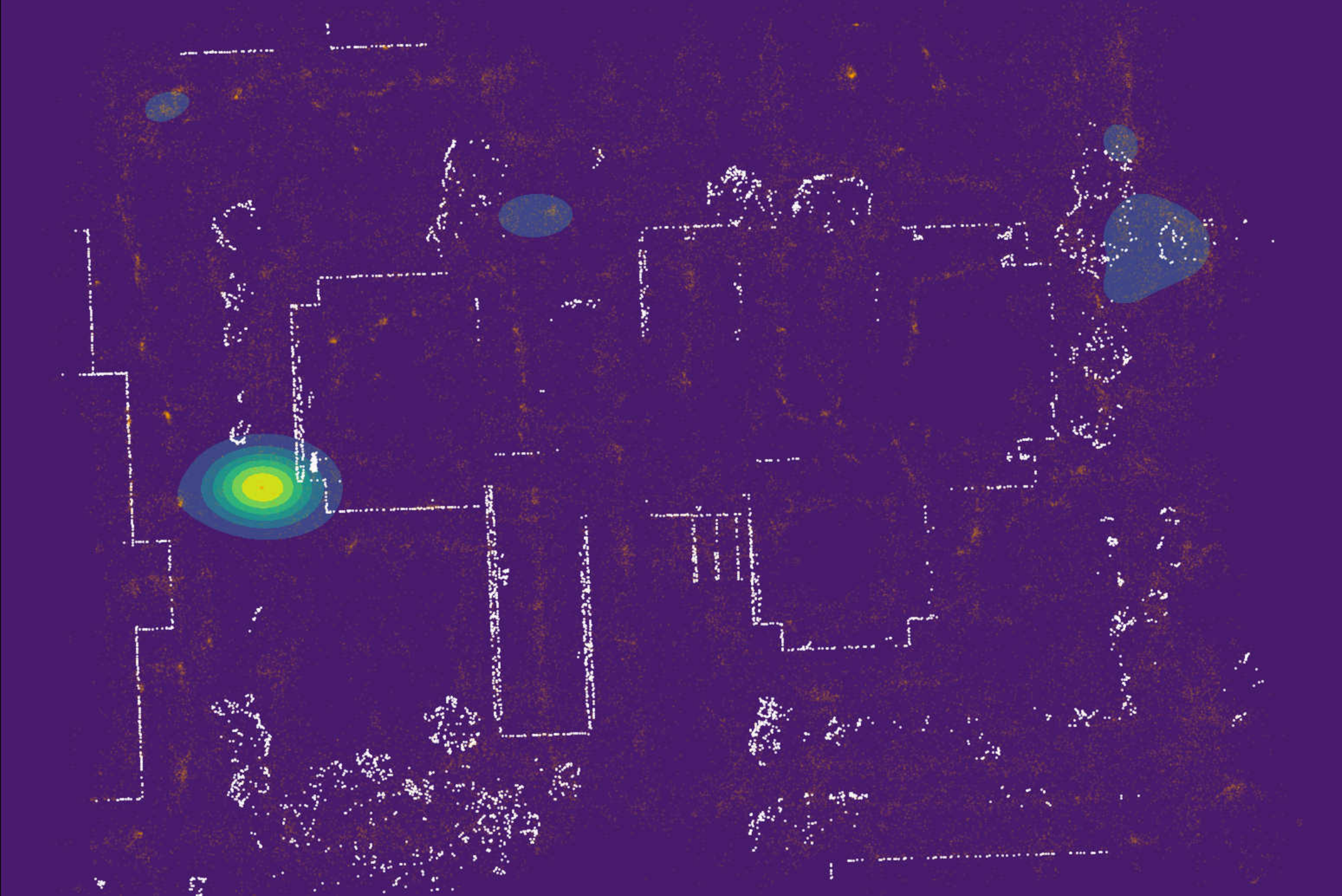}
  \subcaption{}
  \end{minipage}
  \begin{minipage}[tb]{0.45\linewidth}
  \centering
  \includegraphics[width=\linewidth]{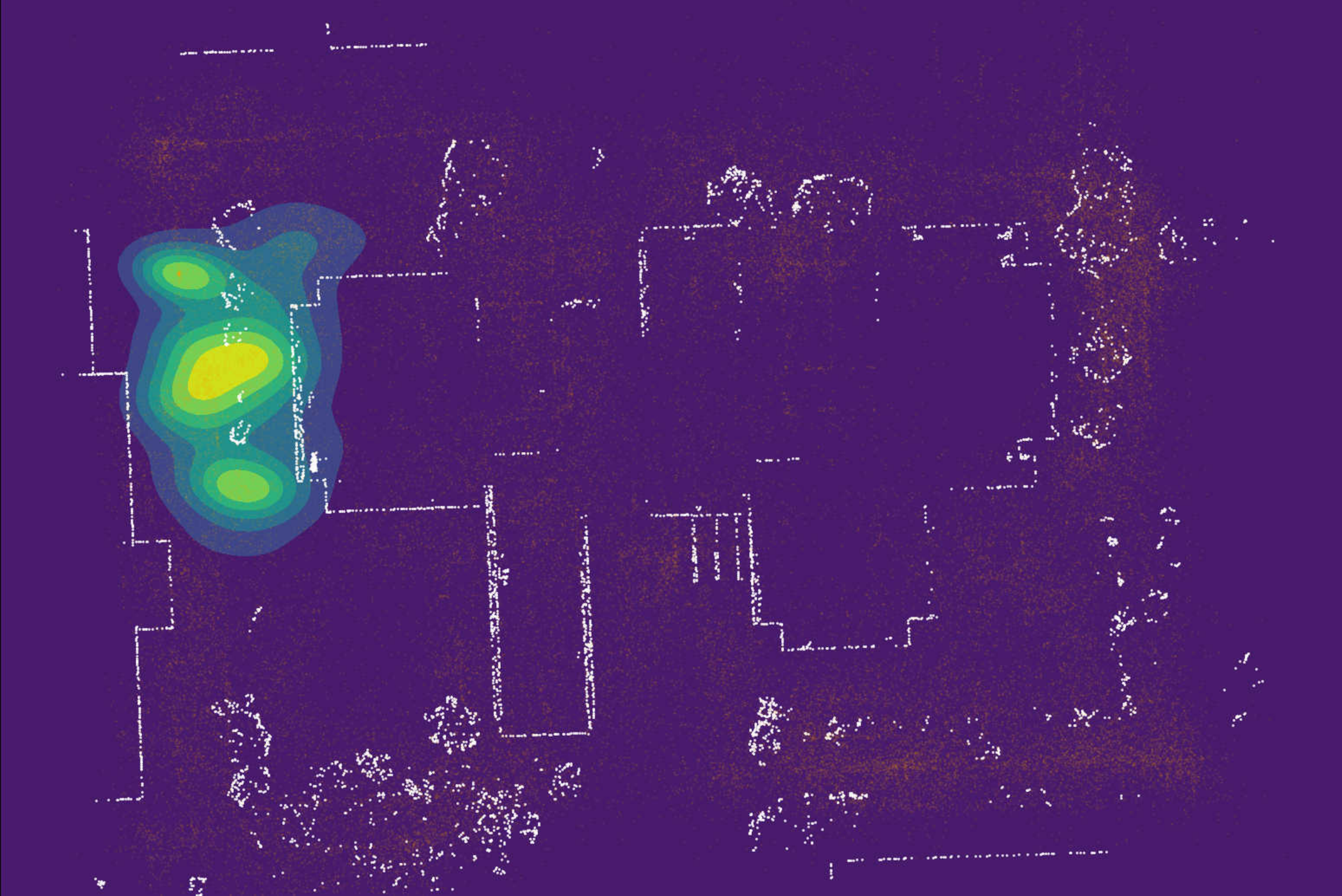}
  \subcaption{}
  \end{minipage}
  \begin{minipage}[tb]{0.45\linewidth}
  \centering
  \includegraphics[width=\linewidth]{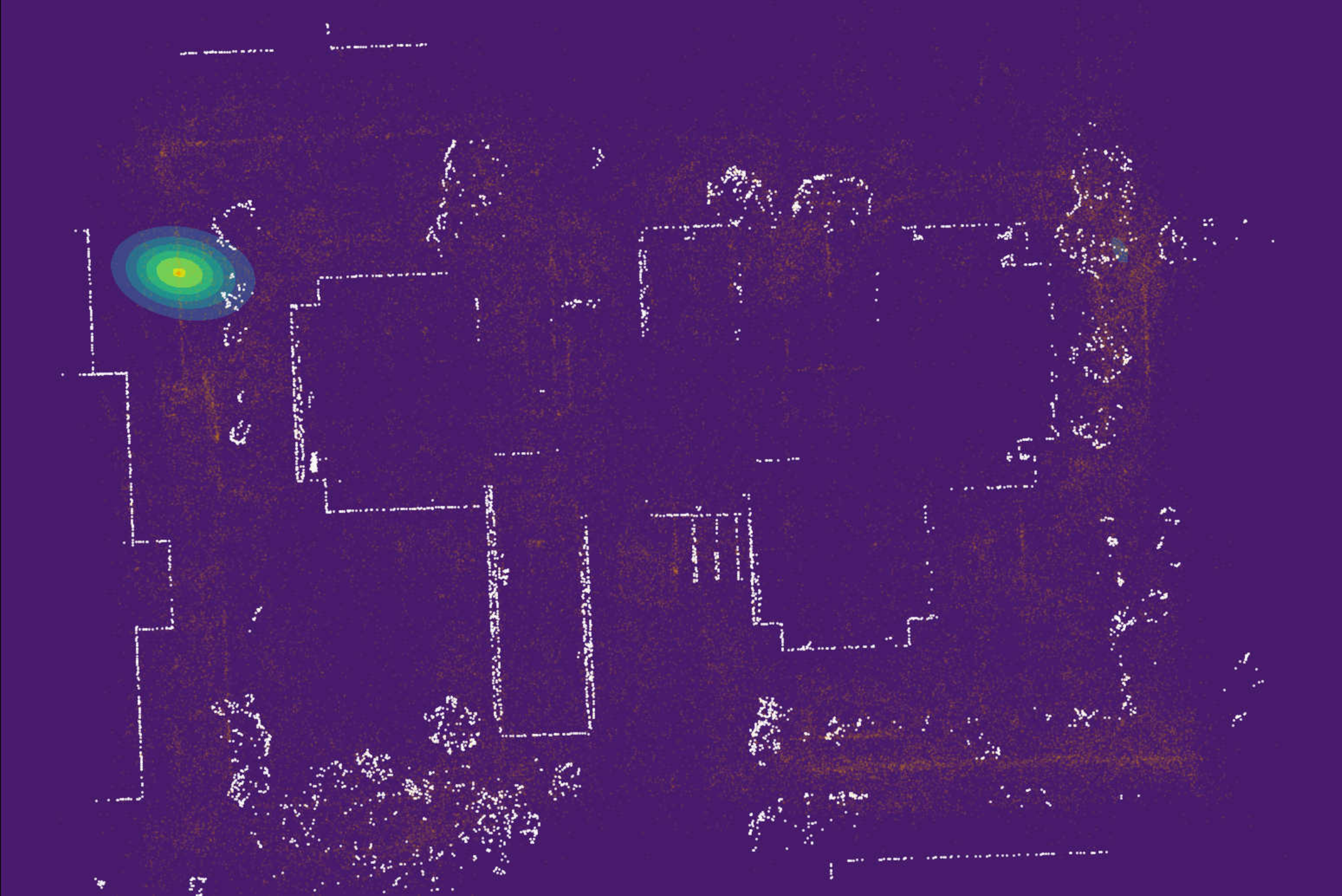}
  \subcaption{}
  \end{minipage}
  \caption{Outdoor experiment results. Uniformly initialized particles quickly converged to the correct position (a, b), and successfully recovered from kidnapping (c, d) during the experiment.}
  \label{fig:outdoor_posterior}
\end{figure}

\begin{figure}[tb]
  \centering
  \includegraphics[width=0.8\linewidth]{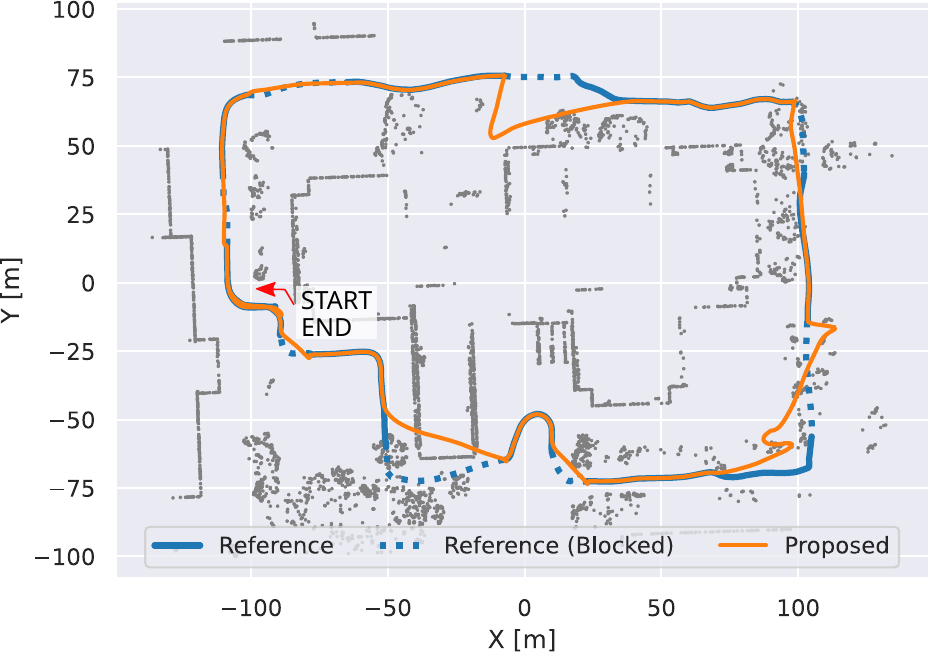}
  \caption{Estimated trajectory for the outdoor experiment. Although the sensor was completely occluded eight times, the proposed method successfully recovered the estimation during the experiment.}
  \label{fig:outdoor_traj}
\end{figure}

{\bf Experimental setting:} We conducted an experiment in the outdoor environment (280 $\times$ 200 $\times$ 30 $\text{m}^3$) shown in Fig. \ref{fig:outdoor}. We used a Livox MID360 to acquire point clouds at 10 Hz. Similar to the indoor experiment, we completely occluded the sensor view eight times during the experiment. As in the indoor experiment, the reference trajectory was obtained through the batch optimization of the registration errors and IMU errors. Note that the scan data were acquired one month after the recording of the map data, and thus there were changes in the dynamic objects (vehicles and pedestrians) and vegetation. The proposed method ran in real-time and took approximately 88.5 ms for each input point cloud.

{\bf Estimation result:} Fig. \ref{fig:outdoor_posterior} shows how the posterior probability distribution changed during the experiment. Starting with an initial uniform distribution (Fig. \ref{fig:outdoor_posterior} (a)), the particles quickly converged to the correct location in a few seconds (Fig. \ref{fig:outdoor_posterior} (b)). Although the posterior distribution diverged during a data interruption (Fig. \ref{fig:outdoor_posterior} (c)), it quickly converged and recovered once point cloud data became available (Fig. \ref{fig:outdoor_posterior} (d)). Fig. \ref{fig:outdoor_traj} shows the estimated trajectory. Although the sensor data were interrupted many times, the proposed method successfully recovered the estimation every time. It required only a few seconds to converge to the correct pose in most of cases. It took longer (15 and 32 s) at two places where the map point cloud was cropped and there was surrounding vegetation. We consider that these map defects and changes prevented the likelihood function from being minimized at the correct location and affected the global localization. 


\section{Conclusion}

This paper presented a particle filter for 6-DoF sensor localization with GPU-accelerated SVGD optimization. A massive number of particles updated in parallel using SVGD on a GPU enabled robust initialization and re-localization without any prior information. 

\balance

\bibliographystyle{IEEEtran}
\bibliography{icra2024}

\begin{thebibliography}{10}
\providecommand{\url}[1]{#1}
\csname url@rmstyle\endcsname
\providecommand{\newblock}{\relax}
\providecommand{\bibinfo}[2]{#2}
\providecommand\BIBentrySTDinterwordspacing{\spaceskip=0pt\relax}
\providecommand\BIBentryALTinterwordstretchfactor{4}
\providecommand\BIBentryALTinterwordspacing{\spaceskip=\fontdimen2\font plus
\BIBentryALTinterwordstretchfactor\fontdimen3\font minus
  \fontdimen4\font\relax}
\providecommand\BIBforeignlanguage[2]{{%
\expandafter\ifx\csname l@#1\endcsname\relax
\typeout{** WARNING: IEEEtran.bst: No hyphenation pattern has been}%
\typeout{** loaded for the language `#1'. Using the pattern for}%
\typeout{** the default language instead.}%
\else
\language=\csname l@#1\endcsname
\fi
#2}}

\bibitem{Koide_2022}
K.~Koide, M.~Yokozuka, S.~Oishi, and A.~Banno, ``Globally consistent and
  tightly coupled 3d {LiDAR} inertial mapping,'' in \emph{{IEEE} International
  Conference on Robotics and Automation}.\hskip 1em plus 0.5em minus
  0.4em\relax {IEEE}, May 2022.

\bibitem{Xu2022}
W.~Xu, Y.~Cai, D.~He, J.~Lin, and F.~Zhang, ``{FAST}-{LIO}2: Fast direct
  {LiDAR}-inertial odometry,'' \emph{{IEEE} Transactions on Robotics}, vol.~38,
  no.~4, pp. 2053--2073, Aug. 2022.

\bibitem{probrobo}
D.~F. Sebastian~Thrun, Wolfram~Burgard, \emph{Probabilistic Robotics}.\hskip
  1em plus 0.5em minus 0.4em\relax The MIT Press, 2005.

\bibitem{Fox_2003}
D.~Fox, ``Adapting the sample size in particle filters through
  {KLD}-sampling,'' \emph{The International Journal of Robotics Research},
  vol.~22, no.~12, pp. 985--1003, Dec. 2003.

\bibitem{Grisetti_2005}
G.~Grisetti, C.~Stachniss, and W.~Burgard, ``Improving grid-based {SLAM} with
  rao-blackwellized particle filters by adaptive proposals and selective
  resampling,'' in \emph{{IEEE} International Conference on Robotics and
  Automation}.\hskip 1em plus 0.5em minus 0.4em\relax {IEEE}, 2005.

\bibitem{Kuang_2023}
H.~Kuang, X.~Chen, T.~Guadagnino, N.~Zimmerman, J.~Behley, and C.~Stachniss,
  ``Ir-mcl: Implicit representation-based online global localization,''
  \emph{IEEE Robotics and Automation Letters}, vol.~8, no.~3, pp. 1627--1634,
  Mar. 2023.

\bibitem{Deng_2021}
X.~Deng, A.~Mousavian, Y.~Xiang, F.~Xia, T.~Bretl, and D.~Fox, ``{PoseRBPF}: A
  rao{\textendash}blackwellized particle filter for 6-d object pose tracking,''
  \emph{{IEEE} Transactions on Robotics}, vol.~37, no.~5, pp. 1328--1342, Oct.
  2021.

\bibitem{Maken_2022}
F.~A. Maken, F.~Ramos, and L.~Ott, ``Stein particle filter for nonlinear,
  non-gaussian state estimation,'' \emph{{IEEE} Robotics and Automation
  Letters}, vol.~7, no.~2, pp. 5421--5428, Apr. 2022.

\bibitem{NIPS2016_b3ba8f1b}
Q.~Liu and D.~Wang, ``Stein variational gradient descent: A general purpose
  bayesian inference algorithm,'' in \emph{Advances in Neural Information
  Processing Systems}, D.~Lee, M.~Sugiyama, U.~Luxburg, I.~Guyon, and
  R.~Garnett, Eds., vol.~29.\hskip 1em plus 0.5em minus 0.4em\relax Curran
  Associates, Inc., 2016.

\bibitem{Chetverikova}
D.~Chetverikov, D.~Svirko, D.~Stepanov, and P.~Krsek, ``The trimmed iterative
  closest point algorithm,'' in \emph{Object recognition supported by user
  interaction for service robots}.\hskip 1em plus 0.5em minus 0.4em\relax
  {IEEE}, 2002, pp. 545--548.

\bibitem{magnusson2009three}
M.~Magnusson, ``The three-dimensional normal-distributions transform: an
  efficient representation for registration, surface analysis, and loop
  detection,'' Ph.D. dissertation, {\"O}rebro universitet, 2009.

\bibitem{Junior_2022}
G.~P.~C. Junior, A.~M.~C. Rezende, V.~R.~F. Miranda, R.~Fernandes, H.~Azpurua,
  A.~A. Neto, G.~Pessin, and G.~M. Freitas, ``{EKF}-{LOAM}: An adaptive fusion
  of {LiDAR} {SLAM} with wheel odometry and inertial data for confined spaces
  with few geometric features,'' \emph{{IEEE} Transactions on Automation
  Science and Engineering}, vol.~19, no.~3, pp. 1458--1471, July 2022.

\bibitem{Saarinen_2013}
J.~Saarinen, H.~Andreasson, T.~Stoyanov, and A.~J. Lilienthal, ``Normal
  distributions transform monte-carlo localization ({NDT}-{MCL}),'' in
  \emph{{IEEE}/{RSJ} International Conference on Intelligent Robots and
  Systems}.\hskip 1em plus 0.5em minus 0.4em\relax {IEEE}, Nov. 2013.

\bibitem{Perez_Grau_2017}
F.~J. Perez-Grau, F.~Caballero, A.~Viguria, and A.~Ollero, ``Multi-sensor
  three-dimensional monte carlo localization for long-term aerial robot
  navigation,'' \emph{International Journal of Advanced Robotic Systems},
  vol.~14, no.~5, Sept. 2017.

\bibitem{Akai_2020}
N.~Akai, T.~Hirayama, and H.~Murase, ``3d monte carlo localization with
  efficient distance field representation for automated driving in dynamic
  environments,'' in \emph{{IEEE} Intelligent Vehicles Symposium}.\hskip 1em
  plus 0.5em minus 0.4em\relax {IEEE}, Oct. 2020.

\bibitem{Maggio_2023}
D.~Maggio, M.~Abate, J.~Shi, C.~Mario, and L.~Carlone, ``Loc-nerf: Monte carlo
  localization using neural radiance fields,'' in \emph{IEEE International
  Conference on Robotics and Automation}.\hskip 1em plus 0.5em minus
  0.4em\relax IEEE, May 2023.

\bibitem{Sun_2020}
H.~Sun, X.~Liu, Q.~Deng, W.~Jiang, S.~Luo, and Y.~Ha, ``Efficient {FPGA}
  implementation of k-nearest-neighbor search algorithm for 3d {LIDAR}
  localization and mapping in smart vehicles,'' \emph{{IEEE} Transactions on
  Circuits and Systems {II}: Express Briefs}, vol.~67, no.~9, pp. 1644--1648,
  Sept. 2020.

\bibitem{Stepanas_2022}
K.~Stepanas, J.~Williams, E.~Hernandez, F.~Ruetz, and T.~Hines, ``{OHM}: {GPU}
  based occupancy map generation,'' \emph{{IEEE} Robotics and Automation
  Letters}, vol.~7, no.~4, pp. 11\,078--11\,085, Oct. 2022.

\bibitem{Peng_2020}
C.~Peng and D.~Weikersdorfer, ``Map as the hidden sensor: Fast odometry-based
  global localization,'' in \emph{{IEEE} International Conference on Robotics
  and Automation}.\hskip 1em plus 0.5em minus 0.4em\relax {IEEE}, May 2020.

\bibitem{segal2009generalized}
A.~Segal, D.~Haehnel, and S.~Thrun, ``Generalized-icp.'' in \emph{Robotics:
  science and systems}, vol.~2, no.~4.\hskip 1em plus 0.5em minus 0.4em\relax
  Seattle, WA, 2009, p. 435.

\bibitem{Datar_2004}
M.~Datar, N.~Immorlica, P.~Indyk, and V.~S. Mirrokni, ``Locality-sensitive
  hashing scheme based on p-stable distributions,'' in \emph{Proceedings of the
  twentieth annual symposium on Computational geometry}.\hskip 1em plus 0.5em
  minus 0.4em\relax {ACM}, June 2004.

\bibitem{teschner2003optimized}
M.~Teschner, B.~Heidelberger, M.~M{\"u}ller, D.~Pomerantes, and M.~H. Gross,
  ``Optimized spatial hashing for collision detection of deformable objects.''
  in \emph{Vmv}, vol.~3, 2003, pp. 47--54.

\bibitem{Arulampalam_2002}
M.~Arulampalam, S.~Maskell, N.~Gordon, and T.~Clapp, ``A tutorial on particle
  filters for online nonlinear/non-gaussian bayesian tracking,'' \emph{{IEEE}
  Transactions on Signal Processing}, vol.~50, no.~2, pp. 174--188, 2002.

\bibitem{Koide_2019}
K.~Koide, J.~Miura, and E.~Menegatti, ``A portable three-dimensional
  {LIDAR}-based system for long-term and wide-area people behavior
  measurement,'' \emph{International Journal of Advanced Robotic Systems},
  vol.~16, no.~2, Mar. 2019.

\bibitem{Zhang_2018}
Z.~Zhang and D.~Scaramuzza, ``A tutorial on quantitative trajectory evaluation
  for visual(-inertial) odometry,'' in \emph{{IEEE}/{RSJ} International
  Conference on Intelligent Robots and Systems}.\hskip 1em plus 0.5em minus
  0.4em\relax {IEEE}, Oct. 2018, pp. 7244--7251.

\end{thebibliography}

\end{document}